# Forecast Network-Wide Traffic States for Multiple Steps Ahead: A Deep Learning Approach Considering Dynamic Non-Local Spatial Correlation and Non-Stationary Temporal Dependency

Xinglei Wang, Xuefeng Guan*, Jun Cao, Na Zhang, Huayi Wu

*State Key Laboratory of Information Engineering in Surveying, Mapping and Remote Sensing, Wuhan University, Wuhan 430079, China*

## Abstract

Obtaining accurate information about future traffic flows of all links in a traffic network is of great importance for traffic management and control applications. This research studies two particular problems in traffic forecasting: (1) capture the dynamic and non-local spatial correlation between traffic links and (2) model the dynamics of temporal dependency for accurate multiple steps ahead predictions. To address these issues, we propose a deep learning framework named *Spatial-Temporal Sequence to Sequence* model (STSeq2Seq). This model builds on sequence to sequence (seq2seq) architecture to capture temporal feature and relies on graph convolution for aggregating spatial information. Moreover, STSeq2Seq defines and constructs pattern-aware adjacency matrices (PAMs) based on pair-wise similarity of the recent traffic patterns on traffic links and integrate it into graph convolution operation. It also deploys a novel seq2sesq architecture which couples a convolutional encoder and a recurrent decoder with attention mechanism for dynamic modeling of long-range dependence between different time steps. We conduct extensive experiments using two publicly-available large-scale traffic datasets and compare STSeq2Seq with other baseline models. The numerical results demonstrate that the proposed model achieves state-of-the-art forecasting performance in terms of various error measures. The ablation study verifies the effectiveness of PAMs in capturing dynamic non-local spatial correlation and the superiority of proposed seq2seq architecture in modeling non-stationary temporal dependency for multiple steps ahead prediction. Furthermore, qualitative analysis is conducted on PAMs as well as the attention weights for model interpretation.

## 1. Introduction

As a key component of Intelligent Transportation Systems (ITS) and Advanced Travel Information Systems (ATIS), accurate traffic forecasting benefits traffic management, urban planning, internet of vehicles and many other applications. For instance, high-precision traffic prediction allows transportation agencies to better control the traffic and reduce congestions. It also enables travelers to foresee traffic conditions on roads and plan appropriate routes accordingly. In this paper, we study network-wide multiple steps ahead traffic forecasting, which aims to predict the future traffic information for the whole road network based on past measurements of traffic on the underlying road network.

Forecast network-wide traffic states for multiple steps ahead is a challenging task due to the non-Euclidean topological structure of traffic network, the stochastic characteristics of the non-stationary

* Corresponding author. Email: guanxuefeng@whu.edu.cn

traffic patterns, and inherent difficulty in multiple steps ahead prediction (Zhang et al., 2019). A collection of researches have been dedicated to solve this task through simultaneously considering the spatial and temporal dependencies of traffic conditions, and deep learning, as an emerging technology, has become the backbone of many newly proposed traffic forecasting models (Li et al., 2018; Yu et al., 2018; Zhang et al., 2019). These deep-learning-based methods adopt graph convolutional neural networks (GCNs) to model the heterogeneous spatial correlations among traffic links, and exploit sequence modeling to capture non-stationary temporal dependency for traffic series. Although dramatic progress has been made in terms of both forecasting accuracy and computational efficiency, we detected following issues that have not been completely solved:

First, spatial modeling had been limited by methods that only consider locally adjacent links. These methods postulate that neighboring links are the most spatially correlated links with the study link and embed their information in forecasting methods as an input (Ermagun and Levinson, 2018b). However, in the real-world, the spatial dependence is often found to exist in a wider range of the traffic networks. For instance, the traffic on two distant roads that both lie within commuting routes tend to exhibit similar patterns even these two roads are far away from each other, thus the traffic information on one road could offer hints for predicting another road's traffic status. Consequently, considering spatial correlation at a local scale could lead to inadequacy in capturing relevant information from distant links, and it could also increase error if the adjacent links have not any spatial effect on the study link. Moreover, since traffic is constantly evolving, the spatial correlations are not static but will change over time. For example, in the morning, the dependency between a residential area and a business center could be strong; whereas in late evening, the relation between these two places might be very weak (Yao et al., 2019). It is necessary to consider the dynamics of spatial correlation. In light of the preceding analysis, we think existing methods can be enhanced by additionally modeling the non-local and dynamic spatial correlations.

Second, existing neural architectures are deficient in modeling non-stationary temporal dependency for multiple steps ahead forecasting. Traffic observations exhibit autocorrelation at the adjacent time intervals and show cyclical patterns due to the effect of human daily routine (Guo et al., 2019b). But the correlation is heterogeneous and the temporal periodicity varies in different time-of-day and day-of-week. Exogenous factors such as weather and holidays makes the temporal dependency more unstable and hard to predict. To model the complex non-stationary temporal dependency for multiple steps ahead forecasting, researchers have either adopted temporal convolution operation (Yu et al., 2018; Guo et al., 2019a; Wu et al., 2019) or RNN-based sequence to sequence (seq2seq) architectures (Li et al., 2018; Do et al., 2019; Hao et al., 2019; Zhang et al., 2019). Temporal convolutions convolve along the temporal dimension, which are highly parallelizable and avoid exploding/vanishing gradients during backpropagation, but they can only generate a single output and thus do not apply to multiple steps ahead scenario. RNN-based structures fit the sequential nature of time series, but require iterative computing and subsequently a longer path when capturing long-range dependencies (Gehring et al., 2017). Integrating both temporal convolution and recurrent architectures could result in a more accurate and efficient model for capturing temporal dependency, but such architectures have not been investigated so far.

To address the aforementioned issues, we propose a novel deep neural architecture for network-wide multiple steps ahead traffic forecasting, which extends the current research on capturing dynamic and non-local spatial correlations among traffic links and deploys an efficient architecture for modeling non-stationary temporal dependency between different time steps. Our proposed framework, namely Spatial

Temporal Sequence to Sequence (STSeq2Seq) model, consists of multiple components that work collaboratively for the prediction task. To be specific, it adopts Multi-Layer Perceptron (MLP) to extract high-dimensional temporal patterns from past traffic series and adaptively constructs Pattern-aware Adjacency Matrices (PAMs) based on the similarity of the extracted patterns. By integrating the pre-defined graph adjacency matrix and the adaptively computed PAMs into graph convolution operations, the model aims to capture both local and dynamic non-local spatial dependencies. As for modeling temporal dependency, STSeq2Seq adopts an encoder-decoder architecture, in which the encoder stacks several spatial-temporal convolutional blocks and a projection layer to extract spatiotemporal features from historical data, whereas the decoder takes an RNN unit as building block to autoregressively decode the extracted features to obtain predictions for multiple steps ahead. Besides, a look-back attention mechanism was designed to account for dynamic temporal dependencies between predicted and historical sequences to improve the model capacity further.

We hypothesize that the proposed PAM supplements the pre-defined adjacency matrix in capturing dynamic and non-local spatial correlation and the proposed seq2seq architecture could improve forecasting accuracy and reduce computation overhead. To corroborate the hypothesis, we selected two publicly-available real-world traffic datasets and conducted comprehensive experiments, in which we compared our framework with state-of-the-art benchmark models, performed ablation study and interpreted the learned PAMs as well as the attention weights.

The remainder of this paper is organized as follows. In section 2, we review spatiotemporal traffic forecasting literature from two perspectives that are closely related to our research objectives: (1) the methods for capturing spatial correlation and (2) temporal dependency modeling for multiple steps ahead forecasting. In Section 3 we formulate the traffic forecasting problem, introduce preliminary methodology, and detail the structure and mathematical formulation of the proposed STSeq2Seq model. In Section 4, the dataset description and experimental settings are given and the results of experiments are analyzed. We conclude the paper in Section 5 by summarizing key contributions and findings and suggesting follow-up research directions.

## 2. Revisiting the spatiotemporal dependency modeling for traffic forecasting

Several recent survey papers have extensively reviewed the literature. Vlahogianni et al. (2014) made a comprehensive review of the short-term traffic forecasting literature up to 2014, and one of the ten under-researched directions pointed out is: “Focus on network level spatio-temporal approaches, fusing modeling and data-driven algorithms”. Ermagun and Levinson (2018b) systematically reviewed articles that embedded the spatial component in traffic forecasting and highlighted that “the spatial correlation between traffic links follows a more sophisticated pattern, which is not simply captured by distance rule”. Following these insights, our work is dedicated to addressing these issues, and we review relevant studies from the following two aspects that connect to our research objectives: (1) methods for capturing spatial correlation; (2) temporal dependency modeling for multiple steps ahead forecasting.

### 2.1. Spatial correlation modeling: from local / static to non-local / dynamic

Modeling spatial correlation is essentially the process of answering the following two questions:

- Which road link have a spatial correlation with the study road link?
- If two road links are correlated, then how much do they correlate with each other?

There have been different perspectives when it comes to answering these two questions, and we introduce them as follows.

*(1) Local V.S. non-local*

To answer the first question, researchers have limited the spatial correlation to be within the $k^{th}$-order neighboring links of the study link, which we refer to as the *local* approach. To decide the magnitude of correlation, they either used binary numbers (i.e. zero and one) to indicate spatial independence and dependence or designed specific rules to calculate the correlation value. For instance, Kamarianakis and Prastacos (2003) examined spatial correlation on both first-order and second-order neighboring links with binary correlation values. Min et al. (2009) measured the spatial correlation as the ratio of traffic flow in each link to the sum of the flow in a T-junction. Li et al. (2018) and Yu et al. (2018) followed the Tobler's First Law of Geography by diminishing the strength of the correlation with distance. Cheng et al. (2011) estimated the correlation as the difference between the average speeds of two links divided by their distance. Although researchers have not come to an agreement about the effective scope of spatial correlation, and has relied on trial and error to decide the magnitude of *k*, *local* approaches are very effective most of the time and has been used in many traffic forecasting methods. On the other hand, some transport researchers took a step forward and start to look into correlation that are not determined by spatial proximity. For example, Pearson correlation coefficient was used as measure of spatial dependency, and results showed both near and distant roads in a traffic network might have high correlation (Sun et al., 2005). Hu et al. (2008) developed the cross-correlation function and found the spatial correlation is not completely dependent on distance. Wu et al. (2019) trained an adaptive weighted adjacency matrix in addition to the pre-determined localized adjacency matrix for capturing the potential spatial correlation beyond the $k^{th}$-order neighboring links. Yu et al. (2019) defined the spatial correlation by calculating similarity distance between two traffic series using Dynamic Time Warping (DTW). Fang et al. (2019) defined a bivariate function with an embedded Gaussian kernel and use it to model both local and non-local spatial correlations. This family of methods assumes that the spatial correlation is not necessarily a function of Euclidean distance in the traffic network, and we refer to them as *non-local* approaches.

*(2) Static V.S. dynamic*

In most of the earlier researches, the spatial correlation was computed once and used throughout the entire prediction process (Sun et al., 2005; Li et al., 2018; Zhang et al., 2019), which implies that spatial correlation is stationary. We refer to this type of methods as *static* approaches. However, the traffic condition on a traffic network is constantly changing, which indicates the necessity of considering the time-varying spatial correlation. Many of the recent studies addressing such issue follow data-driven approaches that dynamically compute the correlation magnitude among traffic links. For example, Yao et al. (2019) adopted a gate mechanism to regulate the spatial correlation based on the changing traffic volume. Guo et al. (2019a) and Do et al. (2019) both defined a spatial attention mechanism, which adjust the distance-based adjacency matrix by the attention score to capture the dynamic local spatial correlation. Likewise, multi-head attention mechanism (Vaswani et al., 2017) was adopted in two similar recent studies (Park et al., 2020; Zheng et al., 2020), in which the spatial correlation was dynamically adapted based on road states.

## 2.2. Temporal dependency modeling: focusing on multiple steps ahead forecasting

It has been acknowledged that forecasting traffic states for multiple steps ahead is more beneficial to practical ITS applications (Vlahogianni et al., 2014; Zhang et al., 2019), for example, it allows transportation agencies more time to optimize the traffic according to the prediction results. In this subsection, we review traffic forecasting methods in terms of temporal dependency modeling, especially for multiple steps ahead forecasting.

Historically, transport researchers have exploited time series analysis, machine learning and deep learning techniques for capturing the complex temporal dependency in relatively long-term forecasting. The autoregressive integrated moving average (ARIMA) model (Ahmed and Cook, 1979), Kalman filtering (Okutani and Stephanedes, 1984), and their variants (Williams and Hoel, 2003; Guo et al., 2014) are the most utilized time series analysis methods and has been widely applied to multiple steps ahead forecasting. Ghosh et al. (2009) introduced a structural time series model to forecast traffic volume for 50 steps ahead. Min and Wynter (2011) utilized time series analysis and provided predictions of speed and volume over 5-min intervals for up to 1 hour in advance. The problem is that simple time series models often place strong stationary assumptions on the data, thus fail to account for highly non-linear temporal dependency, and each step prediction are based on the prior predictions, which lead to the propagation and accumulation of errors. Machine learning models such as k-nearest neighbors (Rice and Van Zwet, 2004), support vector regression (Wu et al., 2004), and artificial neural network (Quek et al., 2006) can model non-linear relations in traffic data, but the relatively shallow architectures limit their power for capturing the complex non-stationary temporal dependencies, especially for long-term forecasting.

Most recently, deep learning models have drawn substantial interests among transport researchers and have been widely applied in modeling temporal dependencies. For example, Yu et al. (2018) proposed a temporal convolution operation and Cui et al. (2019) used a Long Short-Term Memory (LSTM) network (Hochreiter and Schmidhuber, 1997) for capturing the temporal dynamics, respectively. These structures were trained to predict only the next step result and tends to deteriorate in multiple steps ahead prediction. To mitigate this problem, several subsequent methods built on CNN architectures were designed to output the whole prediction sequence directly (Guo et al., 2019a; Wu et al., 2019). These methods require relatively shorter computation time, but ignore the correlations between each time step in the future traffic series. In contrast to convolutional architectures, Li et al. (2018) proposed DCRNN that successfully applied sequence to sequence (seq2seq) architecture (Sutskever et al., 2014) to traffic forecasting, in which both the encoder and decoder are RNN units and multiple steps results were generated in a recursive manner. Many follow-on researches (Do et al., 2019; Hao et al., 2019; Zhang et al., 2019) all adopted similar seq2seq architectures. As a result, this family of models generally takes longer time to train and is less stable than CNN architectures for learning long-term temporal dependencies.

Here we summarize the status quo, and reemphasize the motivation for our research. On the one hand, the problem of capturing the spatial correlation has been studied for decades, yet is far beyond solved. Our goal is to extend the research on it and propose a new effective data-driven approach for capturing the non-local and time-varying factors exist in spatial correlation. On the other hand, the use of deep learning techniques, especially the seq2seq architecture, has substantially improved model's ability to capture temporal dependency, notably under the multiple steps ahead circumstance. However, existing seq2seq architectures that have been used could be improved in terms of both approximation capability and running efficiency, which is the other issue we are addressing in this paper.

## 3. Methodology

In this section, the research problems are formulated and preliminary concepts are introduced, followed by detailed description of the proposed STSeq2Seq model.

### 3.1. Preliminaries

#### 3.1.1. Notations and problem statement

Traffic forecasting can be formulated as a graph modeling problem since the traffic flows are restricted on road networks, which is abstracted as graphs. As Fig. 1(a) shows, a traffic network (e.g. road network or sensor network) is denoted as a directed weighted graph $\mathcal{G} = (\mathcal{V}, \mathcal{E}, \boldsymbol{A})$, where $\mathcal{V}$ is a set of $N$ nodes representing points (roads or sensors) on the road network. $\mathcal{E}$ is a set of edges representing the connectivity among nodes. $\boldsymbol{A} \in \mathbb{R}^{N\times N}$ is the weighted adjacency matrix, in which $\boldsymbol{A}_{i,j}$ indicates the proximity (usually measured by road network distance or topological adjacency) between node $v_i$ and $v_j$. The traffic states of the whole network at time step $t$ is represented as a graph signal $\boldsymbol{X}^t \in \mathbb{R}^{N\times C}$ with $C$ different traffic features (e.g. traffic volume, speed, etc.) on graph $\mathcal{G}$. As Fig. 1(b) shows, given the observations of $P$ historical time steps at $N$ nodes $\boldsymbol{\mathcal{X}} = (\boldsymbol{X}^{t-P+1}, \dots, \boldsymbol{X}^t) \in \mathbb{R}^{P\times N\times C}$, our goal is to predict the traffic states of the next $Q$ time steps for all vertices, denoted as $\boldsymbol{\mathcal{Y}} = (\boldsymbol{Y}^{t+1}, \dots \boldsymbol{Y}^{t+Q}) \in \mathbb{R}^{Q\times N\times C}$.

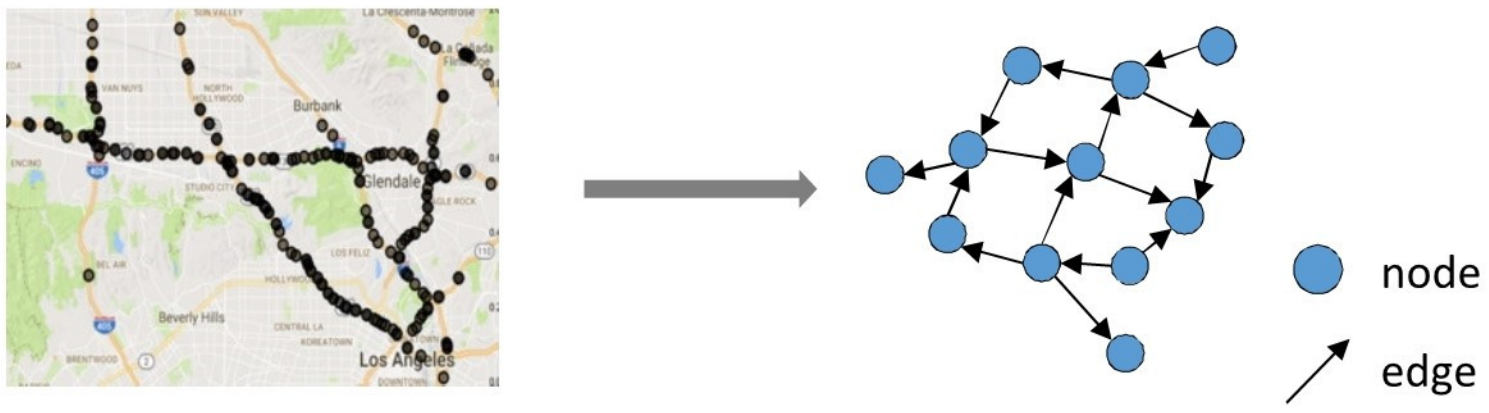

(a) Traffic sensor network abstracted as a graph

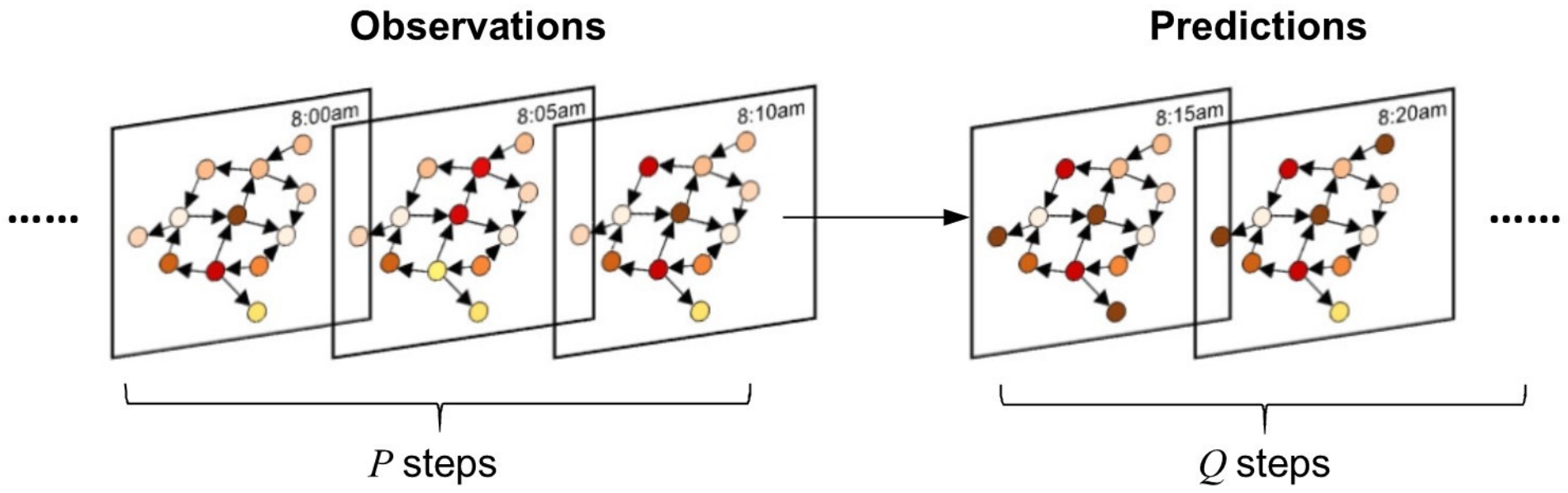

(b) Illustration of network-wide multiple steps ahead traffic forecasting problem

Fig. 1. Illustration of traffic network abstraction and network-wide multiple steps ahead traffic forecasting problem.

#### 3.1.2. Convolution on graphs

A graph convolution operation aggregates node features based on the structure of graphs. Kipf and Welling (2017) proposed a graph convolutional network that simplifies ChebNet (Defferrard et al., 2016) by a first order approximation:

$$\boldsymbol{Y} = \tilde{\boldsymbol{A}}\boldsymbol{X}\boldsymbol{W} \tag{1}$$

where $\tilde{\boldsymbol{A}}$ denotes the normalized adjacency matrix with self-loops, $\boldsymbol{X} \in \mathbb{R}^{N\times D_{in}}$ denote the input graph signals for $N$ nodes with $D_{in}$ features, $\boldsymbol{Y} \in \mathbb{R}^{N\times D_{out}}$ denote the output, and $\boldsymbol{W} \in \mathbb{R}^{D_{in}\times D_{out}}$ denotes the learnable parameter matrix. One limitation of the basic graph convolution is that it only applies to undirected graphs, which violate the directed nature of traffic networks. To facilitate convolution on directed graphs, Li et al. (2018) proposed a forward and backward diffusion process for graph signals with $K$ finite steps. Borrowing from (Wu et al., 2019), diffusion convolution can be generalized into the

form of Eq. (2):

$$\boldsymbol{Y} = \sum_{k=0}^{K} \boldsymbol{M}_f^k \boldsymbol{X} \boldsymbol{W}_{k1} + \sum_{k=0}^{K} \boldsymbol{M}_b^k \boldsymbol{X} \boldsymbol{W}_{k2} \tag{2}$$

where $\boldsymbol{M}^k$ represents the power series of the transition matrix and $K$ is the number of diffusion steps. Forward transition matrix $\boldsymbol{M}_f = \boldsymbol{A}/rowsum(\boldsymbol{A})$ and the backward transition matrix $\boldsymbol{M}_b = \boldsymbol{A}^{\mathrm{T}}/rowsum(\boldsymbol{A}^{\mathrm{T}})$. It has been tested that the bidirectional diffusion convolution can provide forecasting model with the ability and flexibility to capture the influence from both the upstream and the downstream traffic (Li et al., 2018).

### 3.1.3. Sequence to sequence (seq2seq) architecture

Sequence to sequence (seq2seq) architecture has been widely applied in various sequence learning tasks, such as machine translation, speech recognition, multiple steps ahead time series forecasting, etc. This architecture can map arbitrary-length source sequence to arbitrary-length target sequence. As Fig. 2 shows, a typical seq2seq model consists of an encoder and a decoder, both of which are RNN units or their variants, e.g. LSTM and Gated Recurrent Unit (GRU) (Cho et al., 2014). The encoder extracts features from input sequence and encode them into a fixed-size vector representation $c$, which is often the last step hidden state of encoder network. The decoder takes in vector representation $c$ as its initial hidden state and then performing decoding process to progressively produce the output sequence. The process can be written as:

$$h_t = \text{EncoderRNN}(x_t, h_{t-1}) \tag{3}$$

$$s_t = \text{DecoderRNN}(\hat{y}_t, s_{t-1}) \tag{4}$$

$$\hat{y}_t = f(s_t) \tag{5}$$

where $t$ represents certain time step in the encoding or decoding process. $x_t$ is the source input at time $t$, $h_t$ is the encoder's hidden state, $s_t$ is the decoder's hidden state, and the last hidden state of encoder is the initial hidden state of decoder. $f$ is the function (usually a linear layer) that maps decoder's hidden states to the prediction $\hat{y}_t$.

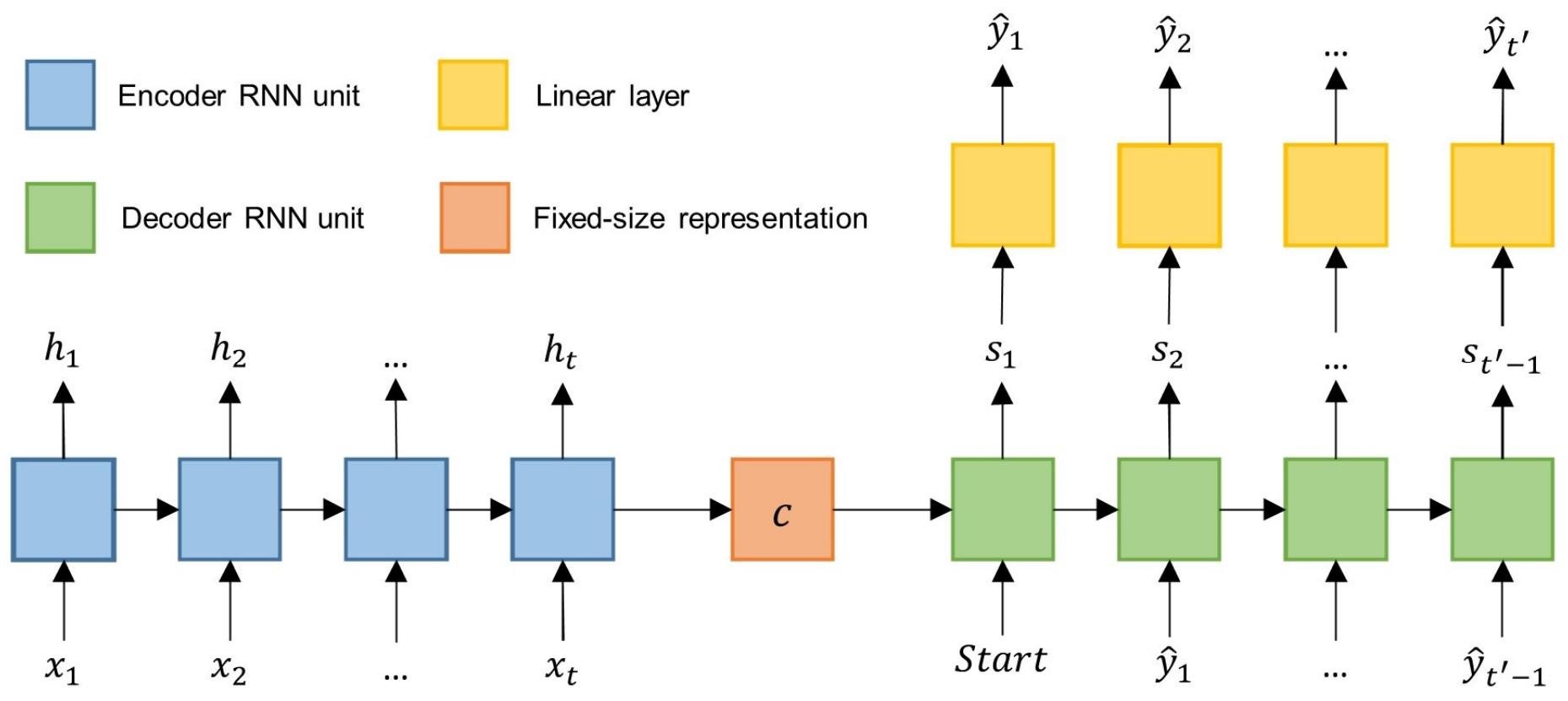

**Fig. 2**. Sequence to sequence architecture.

One downside of the basic seq2seq architecture is that the encoder is trying to cram lots of information into the vector representation. And whilst decoding, the hidden state will need to contain information about the whole of the source sequence, as well as all of the target information decoded so far. To alleviate the information compression, attention mechanism was introduced, enabling the model to pay different attention to specific parts of the input sequence when decoding (Bahdanau et al., 2014; Luong et al., 2015). This relieves the encoder and decoder from squashing all the information into single hidden representation. Attention works by first calculating an attention vector $\boldsymbol{a}$, which has the same length of the source sequence. The attention vector has the property that each element is between 0 and 1, and all elements sum to 1. Then every time step when decoding, a weighted sum of the source hidden states is calculated to get a weighted source vector $w = \sum_i a_i h_i$, which is used as part of the inputs to the decoder RNN unit to make a prediction.

## 3.2. Spatial-temporal sequence to sequence model (STSeq2Seq)

In this subsection, we elaborate on our proposed STSeq2Seq model, which consists of a convolutional encoder to extract spatiotemporal features from the historical traffic data and a recurrent decoder to decode the extracted features and obtain the prediction results.

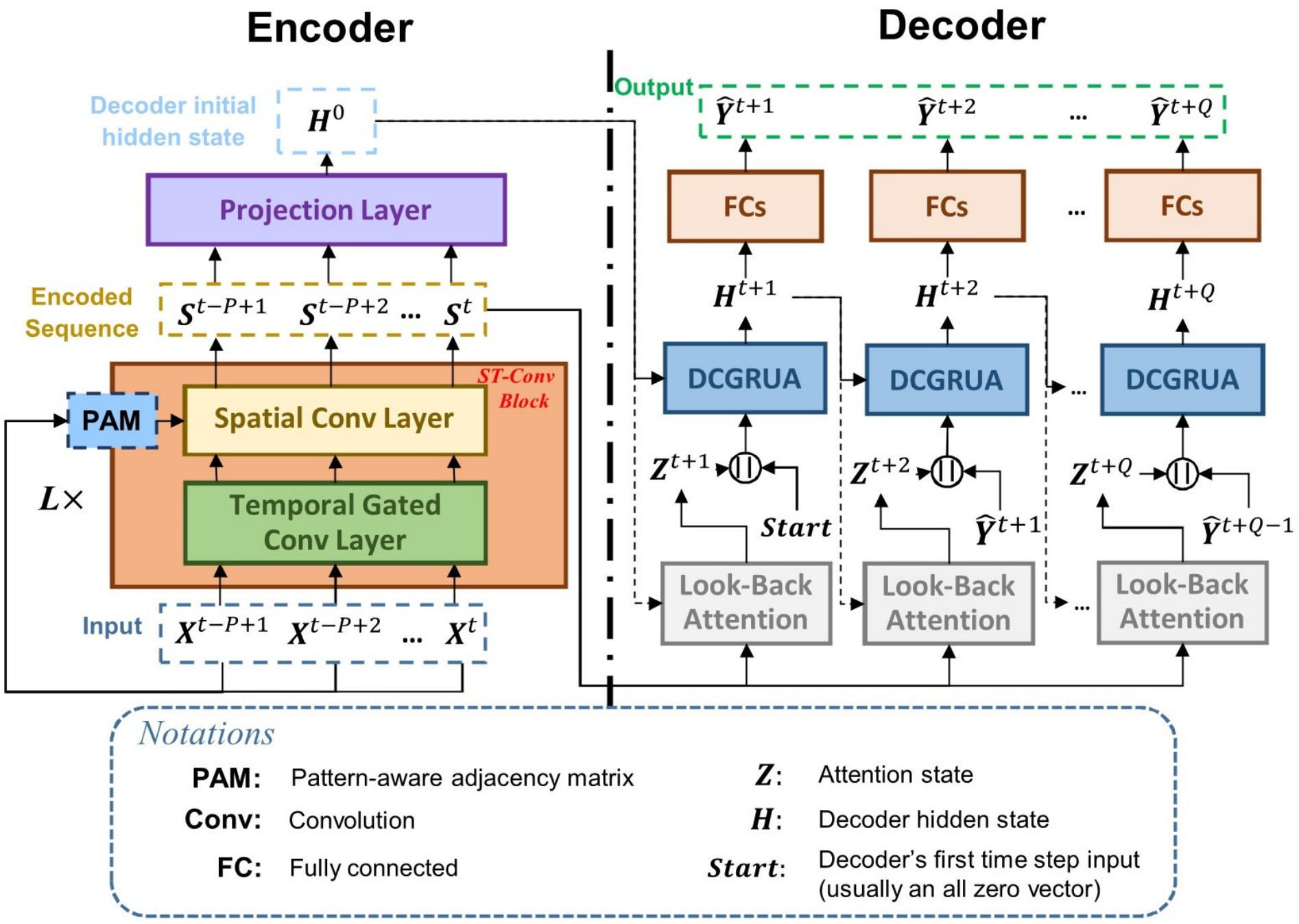

**Fig. 3.** The architecture of STSeq2Seq.

Fig. 3 shows the overall architecture of STSeq2Seq. The encoder stacks spatial and temporal convolution layers to extract hierarchical spatiotemporal features. Pattern-aware adjacency matrix (PAM) is adaptively computed from input historical data and applied in encoder's spatial convolution layer. The decoder is built on RNN units (i.e. Diffusion Convolutional Gated Recurrent Unit (DCGRU) (Li et al., 2018)) with look-back attention mechanism to autoregressively produce predictions for multiple steps. ***Start*** denotes the starting point of decoding stage, and is usually an all zero vector. ‖ denotes the concatenation operation, and the input to the decoder is the concatenation of last step prediction and the attention state $\boldsymbol{Z}$. The details of each of the sub-components of the model are described as follows.

### 3.2.1. Pattern-aware adjacency matrix: considering dynamic non-local spatial correlation

As described in Section 2.1, historical studies have selected spatially relevant links either by distance or by correlation-coefficient analysis (e.g. Pearson correlation), each of which has drawbacks as distance can only reflect local spatial dependency and correlation-coefficient analysis do not capture the non-linear patterns of traffic series. Here we use a different approach which models the non-linear patterns of traffic and account for dynamically changing correlations among roads.

Our approach comprises two steps: non-linear patterns extraction and *pattern-aware adjacency matrix* construction. The core idea is to use the similarity of the real-time temporal pattern as the correlation measure. Intuitively, the more similar the temporal patterns of traffic on two rode links are to each other, the stronger the correlation is between these two roads. However, extracting temporal patterns from time series data is non-trivial as they exhibit high dimensionality, are non-stationary, noisy, and are heavily dependent on the time variable (Längkvist et al., 2014). Various techniques have been applied to model time series, such as restricted Boltzmann machine, autoencoders, and recurrent/convolutional neural networks, etc. More details about time series data modeling can be found in a recent review paper (Längkvist et al., 2014).

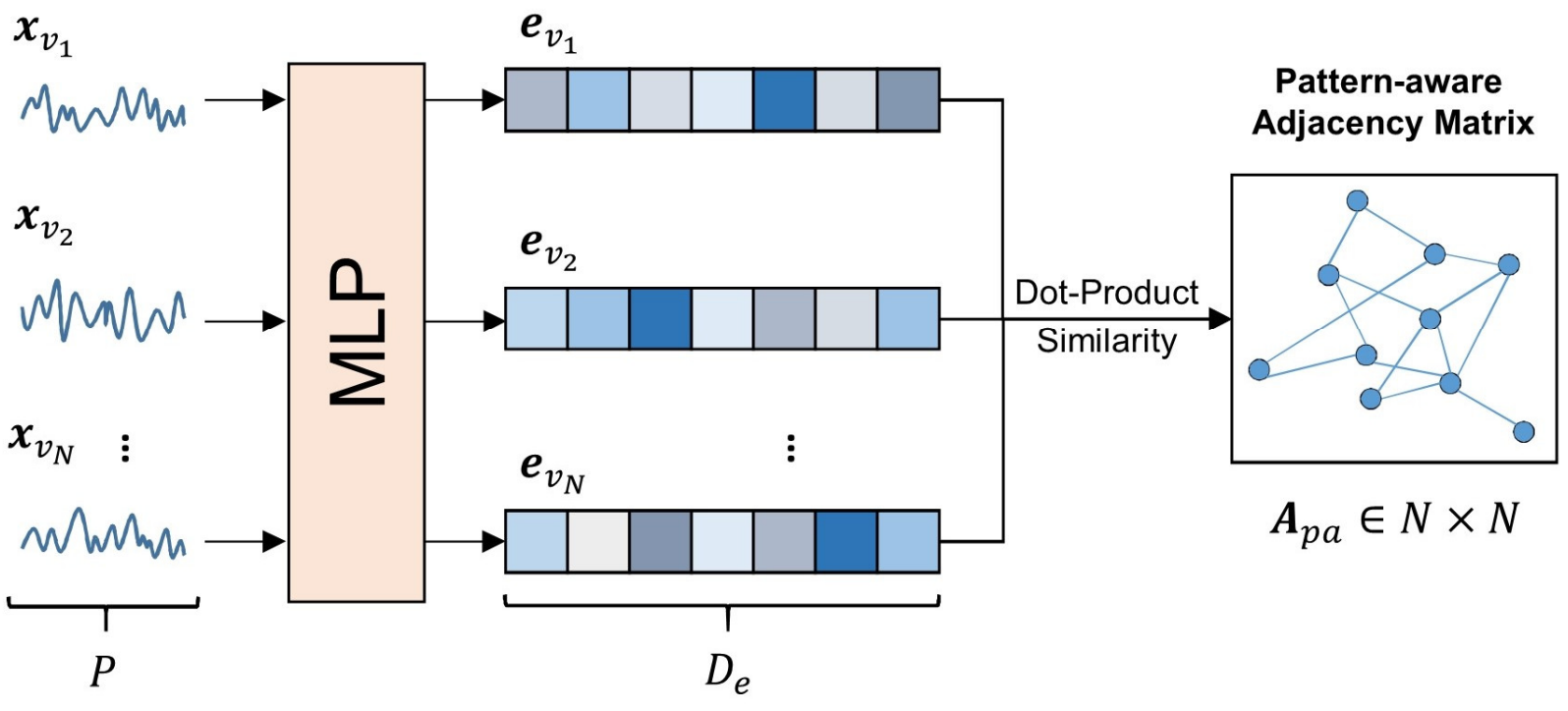

**Fig. 4.** The derivation of pattern-aware adjacency matrix.

Inspired by PointNet (Qi et al., 2017), in which the authors embedded low-dimensional coordinates to high-dimensional vector through Multi-Layer Perceptron (MLP) for better modeling of point cloud data, we propose to use MLP with a single hidden layer to encode the high-dimensional representations of the temporal patterns of historical traffic data. As Fig. 4 shows, given the input time series $\boldsymbol{x}_{v_i}^t \in \mathbb{R}^P$ of a node $\boldsymbol{v}_i$ during period $(t-P+1:t)$, we map it to an embedding $\boldsymbol{e}_{v_i}^t \in \mathbb{R}^{D_e}$, which encode the real-time temporal pattern of the node over the past period. The embedding itself varies with the input data and is used to compute pair-wise similarity between each node. We derive the final correlation weights after applying a Softmax function to normalize the similarity values. The derivation process can be formulized as:

$$\boldsymbol{e}_{v_i}^t = \mathrm{MLP}(\boldsymbol{x}_{v_i}^t) = a(\boldsymbol{b}^{(2)} + \boldsymbol{W}^{(2)}(a(\boldsymbol{b}^{(1)} + \boldsymbol{W}^{(1)}\boldsymbol{x}_{v_i}^t))) \tag{6}$$

$$\boldsymbol{A}_{pa}{}_{i,j}^{t} = \frac{\exp(\boldsymbol{e}_{v_i}^{t\,\mathrm{T}} \cdot \boldsymbol{e}_{v_j}^t)}{\sum_{v \in \mathcal{V}} \exp(\boldsymbol{e}_{v_i}^{t\,\mathrm{T}} \cdot \boldsymbol{e}_{v}^t)} \tag{7}$$

where $\boldsymbol{b}^{(1)} \in \mathbb{R}^{D_e}, \boldsymbol{b}^{(2)} \in \mathbb{R}^{D_e}$ are bias vectors, and $\boldsymbol{W}^{(1)} \in \mathbb{R}^{P \times D_e}, \boldsymbol{W}^{(2)} \in \mathbb{R}^{D_e \times D_e}$ are weight matrices of the MLP with hidden size $D_e$, respectively. $a$ is the activation function (e.g. ReLU), and ${\boldsymbol{A}_{pa}}_{i,j}^{t}$ is the correlation measure between node $v_i$ and $v_j$ during period $(t-P+1:t)$. After computing all the correlation measures, we can get the pattern-aware adjacency matrix $\boldsymbol{A}_{pa}^{t} \in \mathbb{R}^{N\times N}$ within the specified period of time $(t-P+1:t)$.

### 3.2.2. Convolutional encoder for extracting spatial temporal features

As Fig. 3 shows, the encoder is composed of $L$ spatial-temporal convolutional blocks (ST-Conv Blocks) and a projection layer. Each ST-Conv Block contains a temporal gated convolution layer and a spatial convolution layer to extract features from both the temporal and spatial domains. The last block produces the encoded sequence $\boldsymbol{S} = [\boldsymbol{S}^{t-P+1}, \dots, \boldsymbol{S}^{t}] \in \mathbb{R}^{P\times N\times D}$ that encodes hierarchical spatiotemporal features with feature dimension size $D$. A projection layer is designed to fuse the extracted features and generate the representation $\boldsymbol{H}_0 \in \mathbb{R}^{1\times N\times D}$, which is used as the initial hidden state of the decoder.

*(1) Temporal gated convolution layer*

This layer is a 1-D convolution along the temporal dimension followed by gated linear unit (GLU) (Dauphin et al., 2017) as a non-linearity to capture temporal dependency of traffic flows. As Fig. 5 shows, for each node in traffic network $\mathcal{G}$, the temporal convolution explores neighboring time steps of input elements with zero padding so that the temporal dimension size remain unchanged. Given the input of temporal convolution for each node $\boldsymbol{X} \in \mathbb{R}^{P\times D_{in}}$, which is a length-$P$ sequence with $D_{in}$ features, we apply a 1-D convolution kernel $\boldsymbol{\Gamma} \in \mathbb{R}^{K_t\times D_{in}\times 2D_{out}}$ with kernel size $(K_t, 1)$, input size $D_{in}$ and output size $2D_{out}$ to get the output $[\boldsymbol{P}, \boldsymbol{Q}] \in \mathbb{R}^{P\times 2D_{out}}$. $\boldsymbol{P}, \boldsymbol{Q}$ is split in half along the feature dimension and input to GLU. As a result, the temporal gated convolution layer can be written as:

$$\mathrm{TemporalGatedConv}(\boldsymbol{X}) = \mathrm{GLU}(\boldsymbol{\Gamma} * \boldsymbol{X}) = \boldsymbol{P} \odot \boldsymbol{\sigma}(\boldsymbol{Q}) \in \mathbb{R}^{P\times D_{out}} \tag{8}$$

where $\boldsymbol{P}, \boldsymbol{Q}$ are input of gates in GLU respectively, and $\odot$ denotes the element-wise Hadamard product. The sigmoid gate $\sigma(\boldsymbol{Q})$ decides which part of the linear transformation can pass through the gate and contribute to the prediction. Furthermore, the temporal convolution can be generalized to the whole input tensor $\boldsymbol{\mathcal{X}} \in \mathbb{R}^{P\times N\times D_{in}}$ by employing a 2-D convolution kernel $\boldsymbol{\Gamma}' \in \mathbb{R}^{K_t\times 1\times D_{in}\times 2D_{out}}$, denoted as $\boldsymbol{\Gamma}' * \boldsymbol{\mathcal{X}} \in \mathbb{R}^{P\times N\times D_{out}}$.

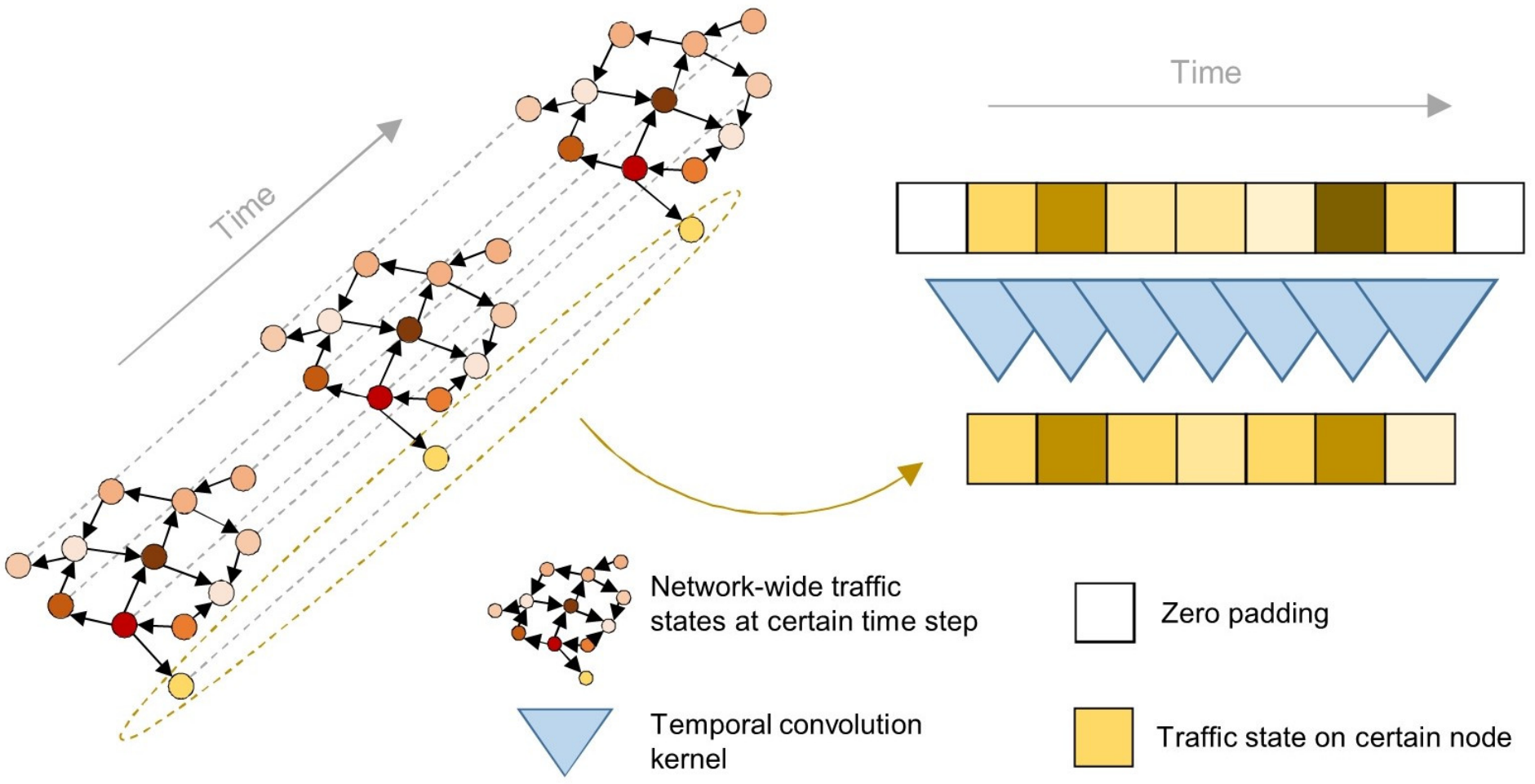

**Fig. 5.** Illustration of temporal convolution.

*(2) Spatial convolution layer*

We propose spatial convolution layer to capture both local and non-local spatial dependencies. As Fig. 6 shows, we use pre-defined weighted adjacency matrix to perform $K$-step diffusion convolution in both forward and backward directions to capture $K$-order localized spatial dependencies, which corresponds to Eq. (2). The PAMs are applied to perform basic graph convolution to model the non-local spatial correlations dynamically, and this process corresponds to Eq. (1). All the convolution results are summed up to obtain the final output graph signals. Formally, given the input of the spatial convolution layer $\boldsymbol{\mathcal{X}} \in \mathbb{R}^{P\times N\times D_{in}}$, we apply spatial convolution to each time slot $\boldsymbol{X}^t \in \mathbb{R}^{N\times D_{in}}$ of the input tensor. The computation can be written as:

$$\text{SpatialConv}(\boldsymbol{X}) = \boldsymbol{A}_{pa}\boldsymbol{X}\boldsymbol{W}_{pa} + \sum_{k=0}^{K} \boldsymbol{M}_f^k \boldsymbol{X}\boldsymbol{W}_{k1} + \sum_{k=0}^{K} \boldsymbol{M}_b^k \boldsymbol{X}\boldsymbol{W}_{k2} \quad (9)$$

where $\boldsymbol{W}_{pa}$ is the learnable parameter matrix for performing convolution with pattern-aware adjacency matrix $\boldsymbol{A}_{pa}$, and all other parameters are the same as described in Eq. (2).

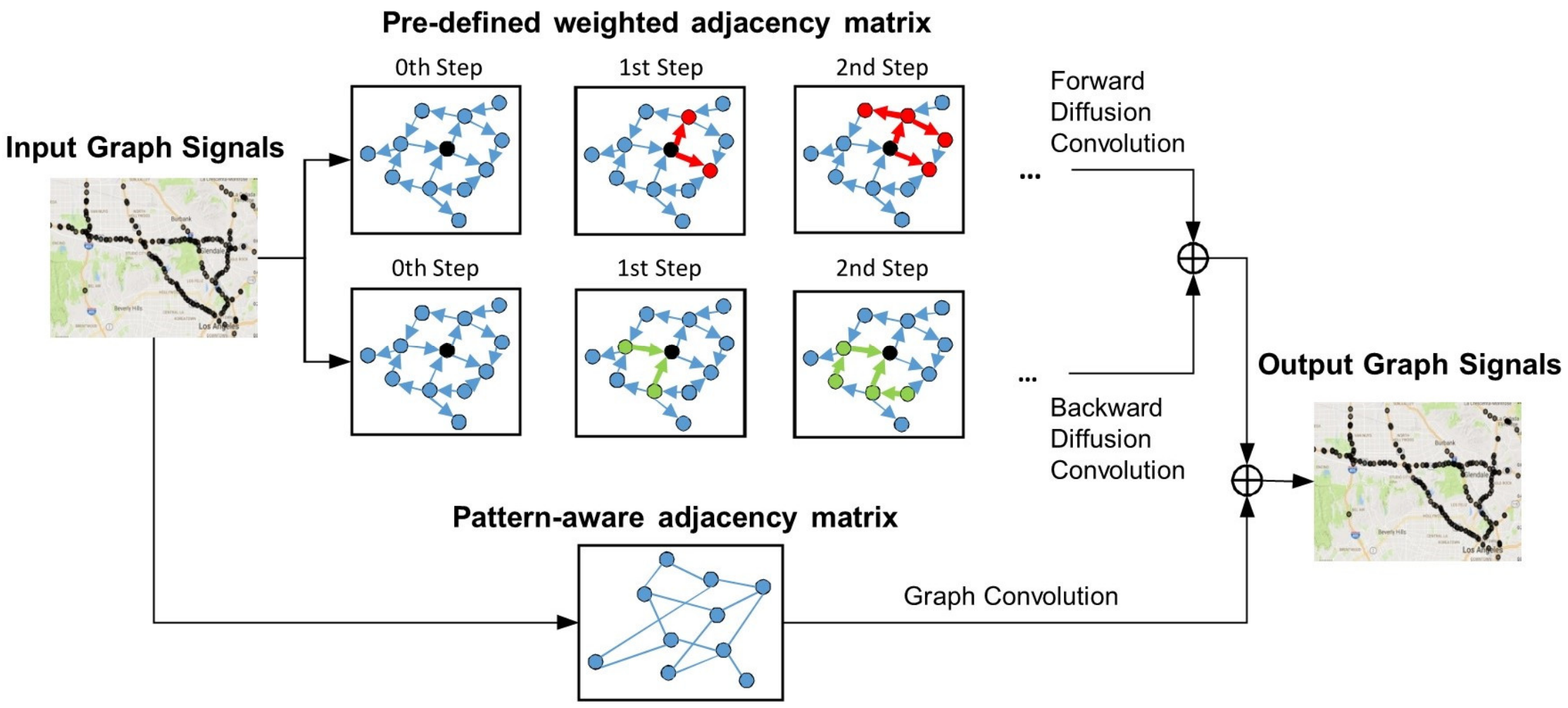

**Fig. 6.** Illustration of spatial convolution layer with demonstration of the diffusion process, where central node is marked black, forward and backward diffusion nodes are marked red and green respectively.

*(3) Projection layer*

The projection layer is essentially a 1-D convolution along the temporal axis with kernel size $K_p = P$, through which the encoded sequence $\boldsymbol{S} \in \mathbb{R}^{P\times N\times D}$ is mapped to $\boldsymbol{H}_0 \in \mathbb{R}^{1\times N\times D}$, which is fed to the decoder as the initial hidden state. This layer has a receptive field spanning the whole input sequence, aggregating the information extracted by previous layers. We argue that such design choice of varying kernel size of the temporal convolution ($K_t = 3, K_P = P$) improves the model capacity as it enables the model to capture spatiotemporal features of the input data at multiple scales.

### 3.2.3. Recurrent decoder for multiple steps ahead prediction

The decoder takes in spatiotemporal features extracted by encoder and produce future traffic series. As Fig. 3 shows, we equip Diffusion Convolutional Gated Recurrent Unit (DCGRU) (Li et al., 2018) with proposed look-back attention mechanism that dynamically model the temporal dependency from historical time steps.

*(1) Look-back attention mechanism*

Historical traffic status at different time steps may have different impacts on the future traffic states. For example, if a traffic accident incurs congestion at 3 p.m., the congestion would have more impact on traffic states during the immediately following time steps than time steps occurring hours later. Considering this phenomenon, we propose a look-back attention mechanism to explicitly model the relevance between the future time steps $i$ $(i = t + 1, \ldots, t + Q)$ and the historical time steps $j$ $(j = t - P + 1, \ldots, t + Q)$. Specifically, when predicting traffic state at time step $i$, we compute an attention state $\boldsymbol{Z}^i$ as the weighted sum of the encoded sequence $\boldsymbol{S}^{t-P+1}, \ldots, \boldsymbol{S}^t$:

$$\boldsymbol{Z}^i = \sum_{j=t-P+1}^{t} \alpha_{ij} \boldsymbol{S}^j \tag{10}$$

$$\alpha_{ij} = \frac{\exp(<\boldsymbol{H}^{i-1}, \boldsymbol{S}^j>_F)}{\sum_{k=t-P+1}^{t} \exp(<\boldsymbol{H}^{i-1}, \boldsymbol{S}^k>_F)}, (i = t + 1, \ldots, t + Q) \tag{11}$$

where $\boldsymbol{H}^{i-1}$ is the hidden state of the decoder at time step $i - 1$ (there is a special case that when $i = t + 1$, we use the initial hidden state $\boldsymbol{H}^0$ for calculation), and $<\cdot,\cdot>_F$ denotes the Frobenius inner product operation. $\alpha_{ij}$ is the attention weight specifying how much the historical traffic state $\boldsymbol{S}^j$ contributes to prediction at time step $i$.

*(2) Diffusion convolutional gated recurrent unit with attention (DCGRUA)*

We propose DCGRUA, which combines DCGRU with look-back attention mechanism to jointly capture the spatial and temporal dependencies in multiple steps ahead forecasting. Specifically, we replace the matrix multiplications in GRU with the diffusion convolution, and feed the context vector at every time step $i$ $(i = t + 1, \ldots, t + Q)$. The computation process of a DCGRUA unit at time $i$ can be formulated as:

$$\boldsymbol{r}^i = \sigma(\mathcal{G}_{\theta_r}[(\boldsymbol{X}^i || \boldsymbol{Z}^i), \boldsymbol{H}^{i-1}] + \boldsymbol{b}_r) \tag{12}$$

$$\boldsymbol{u}^i = \sigma(\mathcal{G}_{\theta_u}[(\boldsymbol{X}^i || \boldsymbol{Z}^i), \boldsymbol{H}^{i-1}] + \boldsymbol{b}_u) \tag{13}$$

$$\boldsymbol{C}^i = tanh(\mathcal{G}_{\theta_c}[(\boldsymbol{X}^i || \boldsymbol{Z}^i), (\boldsymbol{r}^i \odot \boldsymbol{H}^{i-1})] + \boldsymbol{b}_c) \tag{14}$$

$$\boldsymbol{H}^i = \boldsymbol{u}^i \odot \boldsymbol{H}^{i-1} + (1 - \boldsymbol{u}^i) \odot \boldsymbol{C}^i \tag{15}$$

where $\boldsymbol{X}^i$, $\boldsymbol{Z}^i$, $\boldsymbol{H}^i$denote the input graph signal, context vector and the output hidden state at prediction time step $i$ respectively. $\boldsymbol{r}^i$, $\boldsymbol{u}^i$ are the reset gate and update gate at time step $i$, respectively. $\mathcal{G}(\cdot)$ denotes the diffusion convolution as defined in Eq. (2) and $\boldsymbol{\theta}_r$, $\boldsymbol{\theta}_u$, $\boldsymbol{\theta}_C$ are learnable parameters for the corresponding convolution operations. $(\cdot \, || \, \cdot)$ denotes the concatenation operation. Several (typically two) fully connected layers (FCs) were applied at last to map the hidden states from DCGRUA to the final prediction results.

### 3.2.4. Optimization strategy

In the training phase, we defined the loss function as the mean absolute error between the predicted and the ground truth traffic speed sequences and train the model end-to-end via back propagation:

$$\mathcal{L}(\boldsymbol{\theta}) = \frac{1}{|Q|} \sum_{i=t+1}^{t+Q} (\boldsymbol{Y}^i - \widehat{\boldsymbol{Y}}^i) \tag{16}$$

where $\boldsymbol{\theta}$ denotes all the learnable parameter in STSeq2Seq, $\boldsymbol{Y}^i$ and $\widehat{\boldsymbol{Y}}^i$ denote the ground truth and prediction at $i^{th}$ step, respectively.

The standard training strategy for seq2seq model is *teacher forcing*, which feeds ground truths (target sequence) into the decoder for training, and at the testing stage, the predictions generated at last time step are used as input for later time step. However, this could cause degradation of the model due to the discrepancy between the input distributions of training and testing. We mitigate this issue by using *scheduled sampling* (Bengio et al., 2015). Specifically, we feed the model with either the ground truth data with probability $\epsilon_i$ or the prediction generated by the model itself with probability $1-\epsilon_i$ at the $i^{th}$ iteration. During the training phase, $\epsilon_i$ gradually decreases from 1 to 0, which makes the input distribution smoothly change to be consistent with the testing phase.

## 3.3. Insights into the proposed model architectures

Here we provide theoretical analysis of the proposed encoder-decoder structure, and explain the rationale for such design choices in comparison with pure CNN (Yu et al., 2018; Guo et al., 2019a; Wu et al., 2019) and pure RNN (Li et al., 2018; Do et al., 2019; Zhang et al., 2019) architectures.

### 3.3.1. Hierarchical encoding

Our proposed encoder serves as a feature extractor, where the stacked convolutional layers create hierarchical representations over the input sequence as neighboring input time steps interact at lower layers while distant time steps interact at higher layers. Such a structure enriches the extracted information and also provides a shorter path to capture long-range dependencies compared to the chain-structured recurrent networks, since we can obtain a feature representation capturing dependencies within a window of $n$ time steps by performing only $O(n/k)$ convolution operations for kernels of width $k$, as compared to $O(n)$ recurrent computation for RNNs (Gehring et al., 2017). Another feature of convolutional encoder is that it is highly parallelizable and avoids exploding/vanishing gradients during backpropagation, which reduce computation overhead and ease the training of the model.

### 3.3.2. Autoregressive decoding

Our proposed recurrent decoder enables an autoregressive decoding process (Gu et al., 2018), in which each output time step is conditioned on the input historical data as well as previously generated outputs:

$$p_{AR}(\widehat{\boldsymbol{\mathcal{Y}}}|\boldsymbol{\mathcal{X}};\theta) = \prod_{t=1}^{Q} p(\widehat{\boldsymbol{Y}}^t|\widehat{\boldsymbol{Y}}^{0:t-1},\boldsymbol{X}^{1:P};\theta) \tag{17}$$

where $p$ denotes the probability distribution and $\theta$ denotes the learnable parameters of the model. $\widehat{\boldsymbol{Y}}^0$ is the ***Start*** symbol of decoding process. $\boldsymbol{X}^{1:P}$ and $\widehat{\boldsymbol{Y}}^{1:Q}$ denote the historical and predicted sequence, respectively. However, approaches without an autoregressive decoding process (e.g. (Guo et al., 2019a; Wu et al., 2019)) condition the output sequence on the historical sequence only:

$$p_{NAR}(\widehat{\boldsymbol{\mathcal{Y}}}|\boldsymbol{\mathcal{X}};\theta) = \prod_{t=1}^{Q} p(\widehat{\boldsymbol{Y}}^t|\boldsymbol{X}^{1:P};\theta) \tag{18}$$

which exhibits strong conditional independence, incorrectly assuming no correlation exists between each time step of the outputs. By contrast, our recurrent decoder fully considers the temporal correlation in future time series and the chronological order thereof, and is thus preferred.

## 4. Evaluation

### 4.1. Datasets

We verify STSeq2Seq on two publicly-available traffic datasets: METR-LA and PEMS-BAY released by (Li et al., 2018). METR-LA contains 4 months (ranging from Mar 1st 2012 to Jun 30th 2012) of traffic speed data on 207 sensors located on the highway of Los Angeles County. PEMS-BAY contains six months (ranging from Jan 1st 2017 to May 31th 2017) of traffic speed data on 325 sensors in the Bay area. All the speed readings are aggregated into 5 minutes' window. The distribution of sensors are shown in Fig. 7.

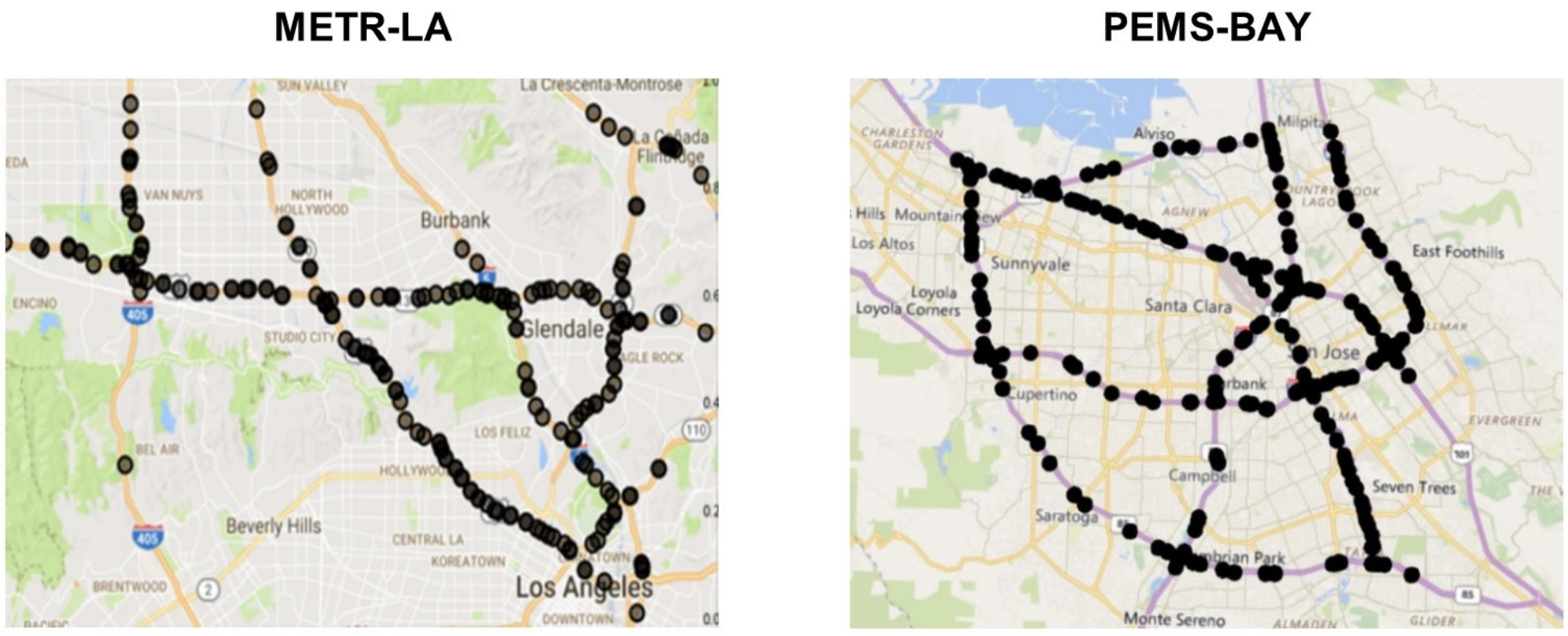

**Fig. 7.** Sensor distribution of the METR-LA and PEMS-BAY dataset.

To compare fairly with other benchmark methods, we strictly followed the data processing procedures outlined in the original papers (Li et al., 2018), which includes: (1) Splitting the dataset in chronological order with the first 70% for training, last 20% for testing and the remaining 10% for validation; (2) Applying Z-score normalization to all the input speed data; (3) Building the weighted adjacency matrix by road network distance with a thresholded Gaussian kernel (Shuman et al., 2013).

$$\boldsymbol{A}_{ij} = \begin{cases} \exp\left(-\frac{\mathrm{dist}(v_i,v_j)^2}{\sigma^2}\right) & \text{if } \mathrm{dist}(v_i,v_j) \le \delta \\ 0 & \text{otherwise} \end{cases} \tag{19}$$

where $\boldsymbol{A}_{ij}$ represents the edge between sensor $v_i$ and sensor $v_j$, $\mathrm{dist}(v_i,v_j)$ denotes the road network distance from sensor $v_i$ to sensor $v_j$. $\sigma$ is the standard deviation of distances and $\delta$ is the threshold. The statistics of the datasets are shown in Table 1. Note that the total speed observations are less than the product of #sensors and #time steps, because there are missing values in the datasets.

**Table 1.** Statistics of METR-LA and PEMS-BAY datasets.

| Dataset | #Sensors | #Edges | #Time Steps | #Observations |
|---|---|---|---|---|
| METR-LA | 207 | 1515 | 34272 | 6,519,002 |
| PEMS-BAY | 325 | 2369 | 52116 | 16,937,179 |

### 4.2. Experimental settings

*(1) Computing environment*

All the experiments are conducted under a computing environment with one Intel Core i7-7700K CPU

@3.20GHz and one NVIDIA RTX 2080Ti GPU. All the deep learning models are implemented with an open source machine learning framework PyTorch (1.2.0 version).

*(2) Baselines*

We compare our model with several baseline models, including time series analysis models (HA, ARIMA), machine learning models (SVR, FNN) and recently published state-of-the-art deep learning models in traffic forecasting domain (LSTM-Seq2Seq, STGCN, DCRNN, Graph WaveNet). The detailed experimental settings for each model are described as follows.

- **HA**: Historical Average, which models the traffic flow as a seasonal process, and uses weighted average of previous seasons as the prediction. For example, the prediction for this Monday is the averaged traffic state of all the past Mondays.
- **ARIMA**: Auto-Regressive Integrated Moving Average model, which is widely used in time series prediction. The model was implemented using the *statsmodel* python package and the orders were set as (3, 0, 1).
- **SVR**: Support Vector Regression, which uses linear support vector machine for the regression task. The model was implemented using *scikit-learn* python package. The penalty term was set as 0.1, kernel type was set as "rbf", and the number of historical observation was set as 5.
- **FNN**: Feed Forward Neural Network with two hidden layers, each contains 256 units and uses L2 regularization.
- **GRU**: seq2seq architecture with attention, in which both the encoder and decoder are fully connected GRU units (Bahdanau et al., 2014). The hidden size of both encoder and decoder GRUs are set as 64.
- **STGCN**: Spatio-Temporal Graph Convolutional Networks, which models spatiotemporal dependencies with graph convolution and 1D temporal convolution (Yu et al., 2018). The open source implementation[†] of the model is used to run the experiment.
- **DCRNN**: Diffusion Convolution Recurrent Neural Network, which combines diffusion convolution with recurrent neural network for highly accurate multiple steps ahead traffic forecasting (Li et al., 2018). The open source implementation[‡] is used to run the experiment. Specifically, the model comprises two layer GRUs with hidden size set to 64.
- **Graph WaveNet**: a model that incorporates graph diffusion convolution into WaveNet (Oord et al., 2016), and also explicitly models the hidden spatial dependencies (Wu et al., 2019). The open source implementation[§] is used in the experiment.

For our model, the number of ST-Conv Blocks $L$ were set to 2 and the *hidden size* of all layers (including MLP, temporal gated convolution layer, spatial convolution layer and DCGRUA) were set to 64. The number of fully connected layers were set as 2. The maximum diffusion step $K$ was set as 1. For the training of all deep learning methods, Adam optimizer was used, learning rate was set as 0.01, and decayed by 0.5 every 10 epochs. All models were trained for 100 epochs with early stopping to prevent model from overfitting.

*(3) Evaluation metrics*

Three commonly used metrics: Mean Absolute Errors (MAE), Mean Absolute Percentage Errors (MAPE) and Root Mean Squared Errors (RMSE) were selected to evaluate the performance of different models. Suppose $\boldsymbol{y} = y_1, \ldots, y_n$ represents the ground truth, $\hat{\boldsymbol{y}} = \hat{y}_1, \ldots, \hat{y}_n$ represents the predicted values, and $Q$ denotes the size of the testing dataset, the calculation of these metrics are:

$$MAE = \frac{1}{Q}\sum_{i=1}^{Q} |y_i - \hat{y}_i| \tag{20}$$

$$RMSE = \sqrt{\frac{1}{Q}\sum_{i=1}^{Q}(y_i - \hat{y}_i)} \tag{21}$$

$$MAPE = \frac{1}{Q}\sum_{i=1}^{Q} \frac{|y_i - \hat{y}_i|}{y_i} \tag{22}$$

[†] https://github.com/VeritasYin/STGCN_IJCAI-18
[‡] https://github.com/liyaguang/DCRNN
[§] https://github.com/nnzhan/Graph-WaveNet

## 4.3. Results and analyses

### 4.3.1. Forecasting performance comparison with baselines

Table 2 presents the forecasting error of STSeq2Seq and other baseline models for 15-min, 30-min and 60-min ahead on the test set. We can observe from the results that:

(1) The performance of HA is invariant to the increases in the forecasting horizon because it simply sets the value for each predicted time interval as the average traffic speed in the predicted time interval for all previous days.

**Table 2.** Performance comparison of STSeq2Seq and other baselines under different forecasting horizons. The best results are marked in bold.

| Data | Models | 15 min | | | 30 min | | | 60 min | | |
|---|---|---|---|---|---|---|---|---|---|---|
| | | MAE | RMSE | MAPE | MAE | RMSE | MAPE | MAE | RMSE | MAPE |
| METR-LA | HA | 4.16 | 7.80 | 13.00% | 4.16 | 7.80 | 13.00% | 4.16 | 7.80 | 13.00% |
| | ARIMA | 3.99 | 8.21 | 9.60% | 5.15 | 10.45 | 12.70% | 6.90 | 13.23 | 17.40% |
| | SVR | 3.99 | 8.45 | 9.30% | 5.05 | 10.87 | 12.10% | 6.72 | 13.76 | 16.70% |
| | FNN | 3.99 | 7.94 | 9.90% | 4.23 | 8.17 | 12.90% | 4.49 | 8.69 | 13.20% |
| | GRU | 2.93 | 5.78 | 7.78% | 3.41 | 6.98 | 9.66% | 4.04 | 8.37 | 12.00% |
| | STGCN | 2.88 | 5.74 | 7.62% | 3.47 | 7.24 | 9.57% | 4.59 | 9.40 | 12.70% |
| | DCRNN | 2.66 | 5.16 | 6.83% | 3.06 | 6.28 | 8.36% | 3.54 | 7.46 | 10.24% |
| | Graph WaveNet | 2.69 | 5.15 | 6.90% | 3.07 | 6.22 | 8.37% | 3.53 | 7.37 | 10.01% |
| | STSeq2Seq | **2.64** | **5.10** | 6.72% | **3.02** | **6.18** | **8.16%** | **3.47** | **7.36** | **9.96%** |
| PEMS-BAY | HA | 2.88 | 5.59 | 6.80% | 2.88 | 5.59 | 6.80% | 2.88 | 5.59 | 6.80% |
| | ARIMA | 1.62 | 3.30 | 3.50% | 2.33 | 4.76 | 5.40% | 3.38 | 6.50 | 8.30% |
| | SVR | 1.85 | 3.59 | 3.80% | 2.48 | 5.18 | 5.50% | 3.28 | 7.08 | 8.00% |
| | FNN | 2.20 | 4.42 | 5.19% | 2.30 | 4.63 | 5.43% | 2.46 | 4.98 | 5.89% |
| | GRU | 1.38 | 2.95 | 2.88% | 1.77 | 4.02 | 4.00% | 2.21 | 5.04 | 5.31% |
| | STGCN | 1.36 | 2.96 | 2.90% | 1.81 | 4.27 | 4.17% | 2.49 | 5.69 | 5.79% |
| | DCRNN | 1.31 | 2.75 | 2.73% | 1.64 | 3.74 | 3.69% | 1.96 | 4.55 | 4.62% |
| | Graph WaveNet | 1.30 | 2.74 | 2.73% | 1.63 | **3.70** | 3.67% | 1.95 | 4.52 | 4.53% |
| | STSeq2Seq | **1.30** | **2.73** | **2.72%** | **1.62** | 3.72 | **3.61%** | **1.92** | **4.48** | **4.42%** |

(2) ARIMA yields better results than HA and machine learning models (SVR, FNN) on the 15 min

horizon, but it degrades rapidly as the prediction horizon increases, which indicates the limited ability to capture non-linear traffic patterns in the long-term.

(3) Deep learning based methods make better predictions than machine learning models, especially those which jointly model both spatial and temporal dependencies.

(4) STGCN considers both spatial and temporal dependencies, but is outperformed by simple GRU on 30-min and 60-min ahead, because it is designed for one-step prediction and does not suit multiple steps ahead prediction scenario.

(5) DCRNN integrates diffusion convolution into GRU and adopts seq2seq architecture, whereas Graph WaveNet benefits from dilated convolution and explicit modeling of hidden spatial dependencies, both of them achieved extremely high forecasting accuracy.

(6) STSeq2Seq outperforms all baselines on the METR-LA dataset, and achieves forecasting accuracy that is on par with state-of-the-art models (i.e. DCRNN and Graph WaveNet) on PEMS-BAY dataset. We think this is because our model not only captures dynamic non-local spatial dependencies but also considers the correlation among predicted time steps through the autoregressive decoding process.

To better illustrate the model's forecasting ability, we randomly select sensor 0 and visualize in Fig. 8 one-day forecasting results of STSeq2Seq and GRU under 15-min, 30-min and 60-min horizons, respectively. Looking at the results at around 16:00, June 04, we can see that STSeq2Seq predicted abrupt changes in the traffic more accurately than the baseline method, GRU. And this superiority becomes more distinct as the forecasting horizon increases. Moreover, STSeq2Seq tends to generate smoother forecasts when the traffic status oscillates seriously, this can be observed throughout the prediction, especially around 18:00, June 04. We believe that this is because STSeq2Seq considers the spatial dependency, and is able to exploit the information in highly correlated sensors for more accurate forecasting.

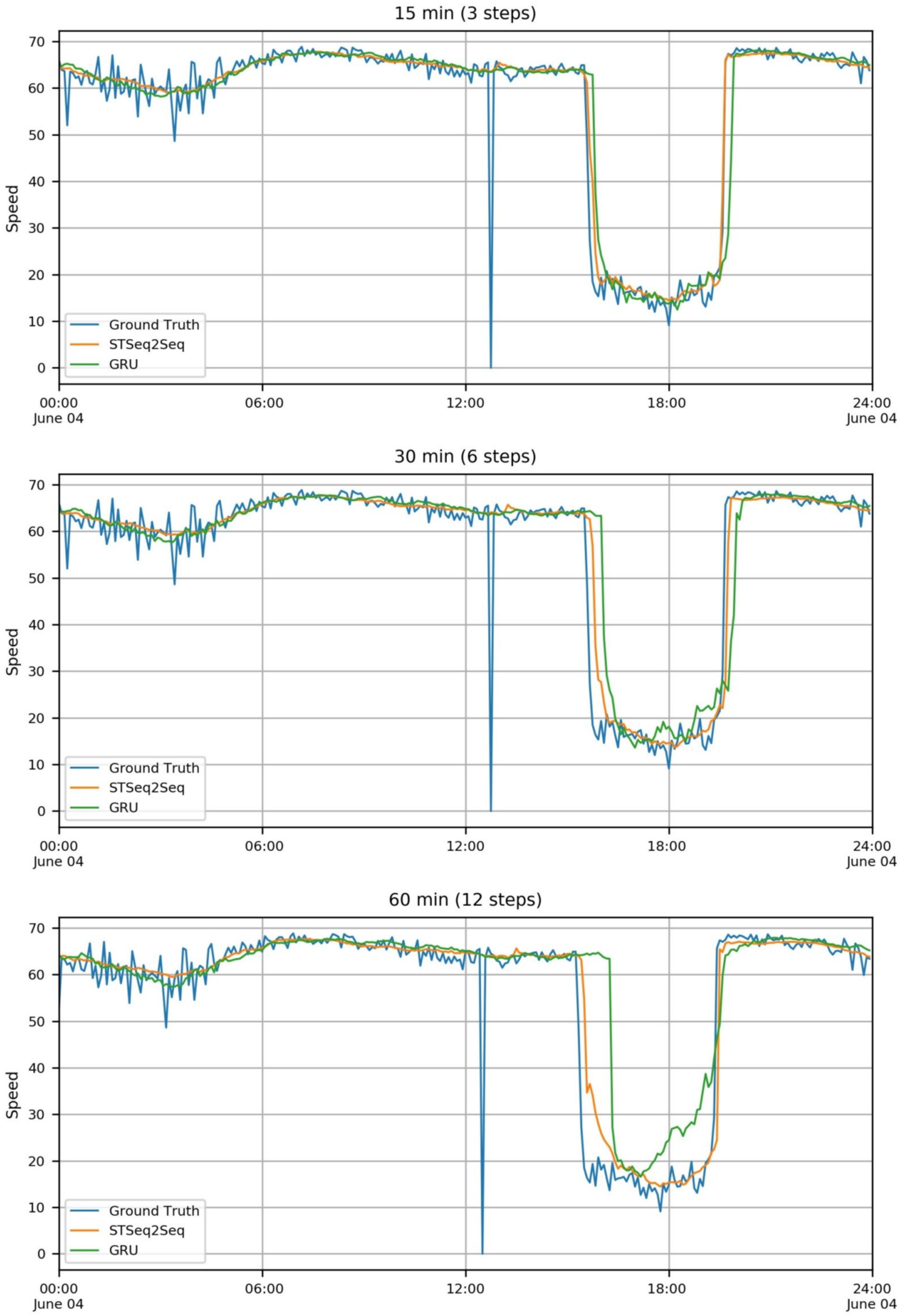

**Fig. 8.** Multiple steps ahead prediction values of STSeq2Seq and GRU on sensor 0.

### 4.3.2. Sensitivity analysis

To investigate how different hyperparameters affect the model performance and also to choose the optimal settings of hyperparameters for our model, we conduct sensitivity analysis on two main hyperparameters: the max diffusion steps $K$ and the hidden size of all layers. $K$ roughly corresponds to the size of spatial convolutions' reception fields while the hidden size corresponds to the number of hidden features. We did experiments using the STSeq2Seq model without applying pattern-aware adjacency matrix for the purpose of studying the effect of pure spatial proximity in modeling spatial dependency.

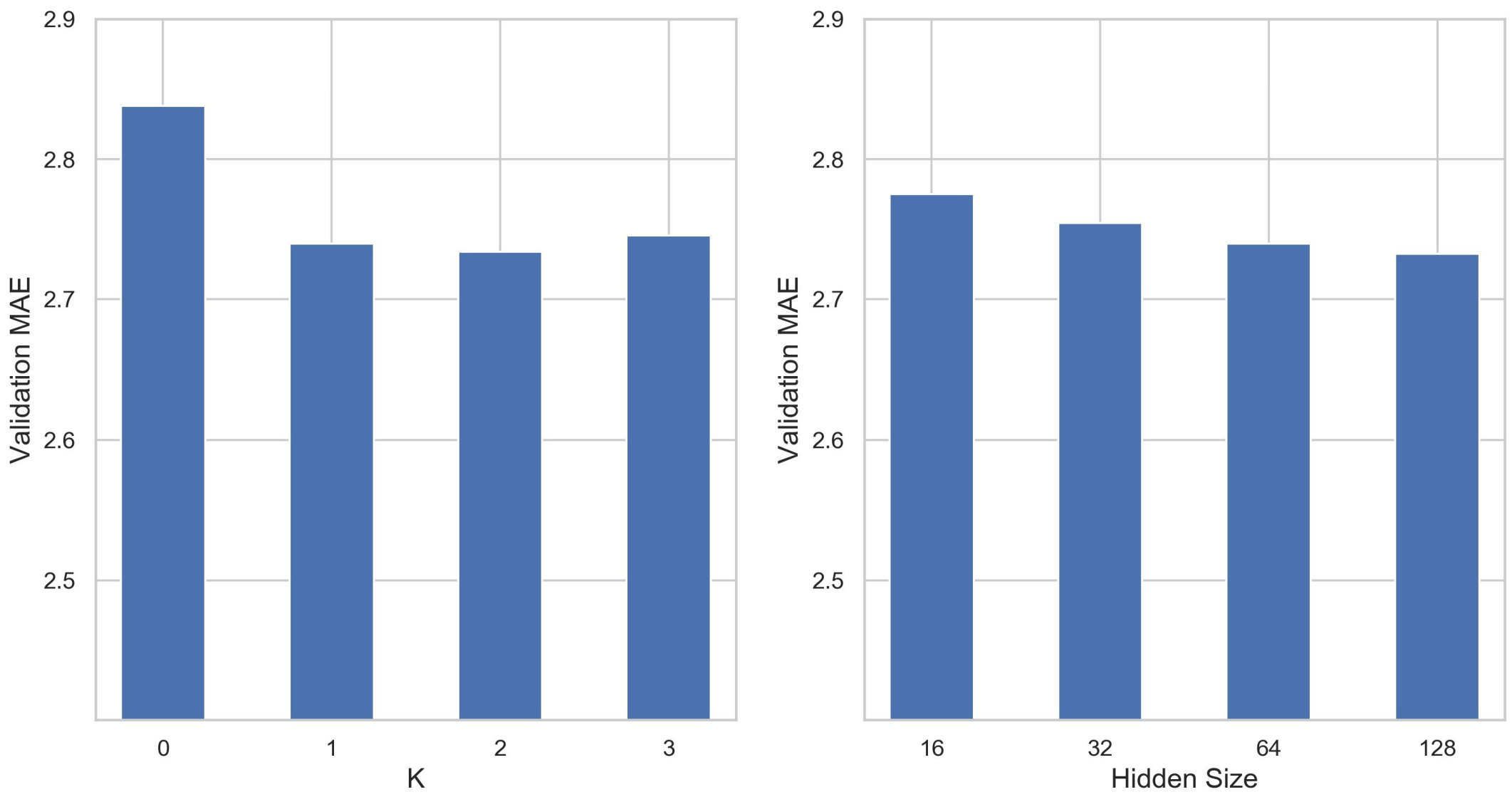

**Fig. 9.** Effects of K and the hidden size in each layer of STSeq2Seq.

Fig. 9 shows the effects of different parameters on the METR-LA dataset. It can be observed that with the increase of $K$, the error on the validation set first drops quickly, and then goes up slightly. While the error keeps going down with increase of the hidden size. We think that larger $K$ enables the model to capture broader range of spatial dependency, but incorporating high-order neighboring links' information may harm the model if the links are not actually correlated with the study link. As for the hidden size, larger hidden size improves the model capacity but it also raises learning complexity and requires more training data. In addition, increasing $K$ or hidden size significantly increases the running time. Therefore, we chose $K$=1 and hidden size=64 for our model as a trade-off between forecasting performance and running efficiency.

### 4.3.3. Effect of spatial dependency modeling

To investigate the effect of spatial dependency modeling, we evaluated STSeq2Seq with different adjacency matrix settings described as follows. *Identity* replaces the transition matrices in the diffusion convolution with identity matrices and exclude pattern-aware adjacency matrix, which assumes that the traffic state of each sensor has nothing to do with the others, but only itself. *Geo-only* excludes the pattern-aware matrix and only uses the weighted adjacency matrix, which rely purely on geographical information for determining spatial correlation (Eq. (2)). *Self-adpt* is similar to *STSeq2Seq* except that it replaces the pattern-aware adjacency matrix with self-adaptive adjacency matrix, which is trained to capture hidden spatial correlation that cannot be reflected by weighted adjacency matrix (Wu et al., 2019). *STSeq2Seq* is the complete form of the proposed model, which uses both the weighted adjacency matrix and pattern-aware adjacency matrix (Eq. (9)).

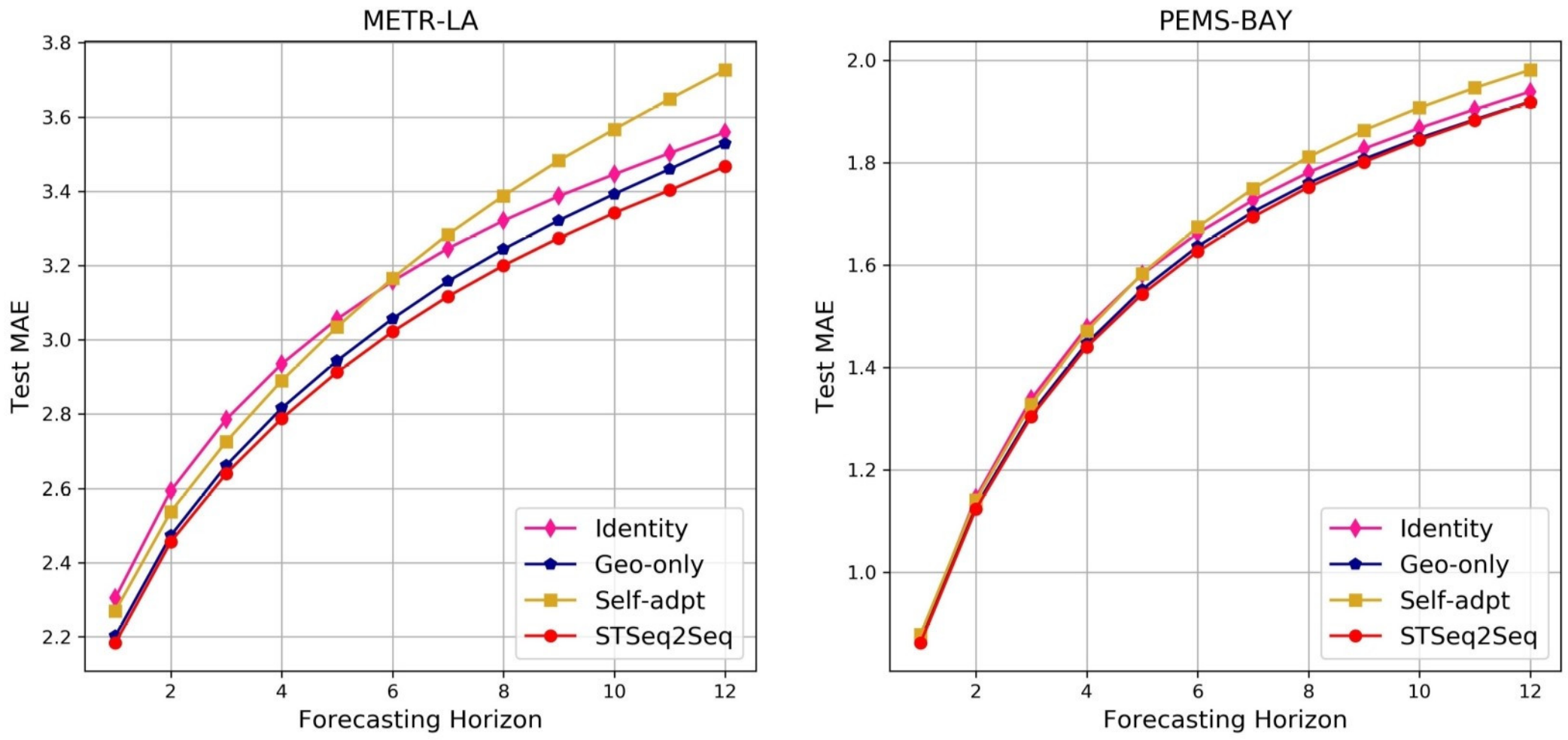

**Fig. 10.** Test MAE under different configurations of adjacency matrix.

Fig. 10 shows the comparative MAE results for different forecasting horizons on METR-LA and PEMS-BAY datasets. We can observe that the MAE of *Geo-only* is consistently smaller than *Identity* for all forecasting horizons, which indicates the necessity and importance of considering spatial factors for traffic prediction. *Self-adpt*, except for the 1~6 steps on METR-LA dataset, obtained lowest forecasting accuracy among all the variant models. We think that the self-adaptive adjacency matrix used in *Self-adpt* did not learn useful information, and the reason for that may be it was fixed throughout the whole forecasting process, and did not capture the dynamic spatial correlations among nodes. *STSeq2Seq* achieves the highest forecasting accuracy on both datasets, verifying the effectiveness of incorporating pattern-aware adjacency matrix, which enables the model to exploit the dynamic traffic patterns and learn useful non-local and time-varying spatial correlations for traffic forecasting. One thing to be noted is that *STSeq2Seq* did not exceed *Geo-only* much on PEMS-BAY dataset, which implies less dynamicity and non-locality of spatial correlation due to the relatively stable traffic condition in Bay Area (Li et al., 2018).

We further look into different types of adjacency matrix learned from the METR-LA dataset. In particular, the heatmaps of the weighted adjacency matrix, the learned self-adaptive adjacency matrix and the pattern-aware adjacency matrices obtained within two typical time intervals were visualized in Fig. 11. Note that only the first 50 nodes were visualized for a better viewing. Our findings based on the visualization results are stated as follows:

(1) As depicted in Fig. 11(a), the weighted adjacency matrix exhibits highest values on the diagonal, since it is based on network distance.

(2) As shown in Fig. 11(b), the self-adaptive adjacency matrix is sparse and fixed. Although it differs greatly from the weighted adjacency matrix, it is hard to interpret the information contained.

(3) As visualized in Fig. 11(c) and (d), the learned pattern-aware matrices are more dense than weighted adjacency matrix, which suggests the capture of non-local connections among nodes. They also vary significantly although they are within two consecutive periods of time, verifying the dynamicity of the learned spatial correlations.

Taking a closer look at the magnitude of the learned correlation in Fig. 11(d), we can observe that the correlation between sensor 9 and 14 was much stronger than others even though they are far away from each other (see Fig. 12). This suggests that our method went beyond the spatial proximity and succeeded in capturing the non-local spatial correlations. We further visualize the traffic time series of the sensors within corresponding time intervals in Fig. 13. Comparing Fig. 11(d) and Fig. 13(right), we can see that the degree of similarity in temporal patterns well matches the magnitude of the learned correlations. For example, the traffic series of sensor 9 and 14 drawn by solid lines exhibit similar patterns, and the correlation between them is significantly higher than others, which, to some extent, verify that the pattern-aware adjacency matrix truly learned useful information about the temporal pattern. And since the traffic series within interval 04:20-05:20 show unclear relations (see Fig. 13(left)), the model assigned extremely low values (an order of magnitude lower than interval 05:20-06:20) to the pattern-aware adjacency matrix. In such case, the model may rely mostly on the weighted adjacency matrix for capturing the spatial dependency.

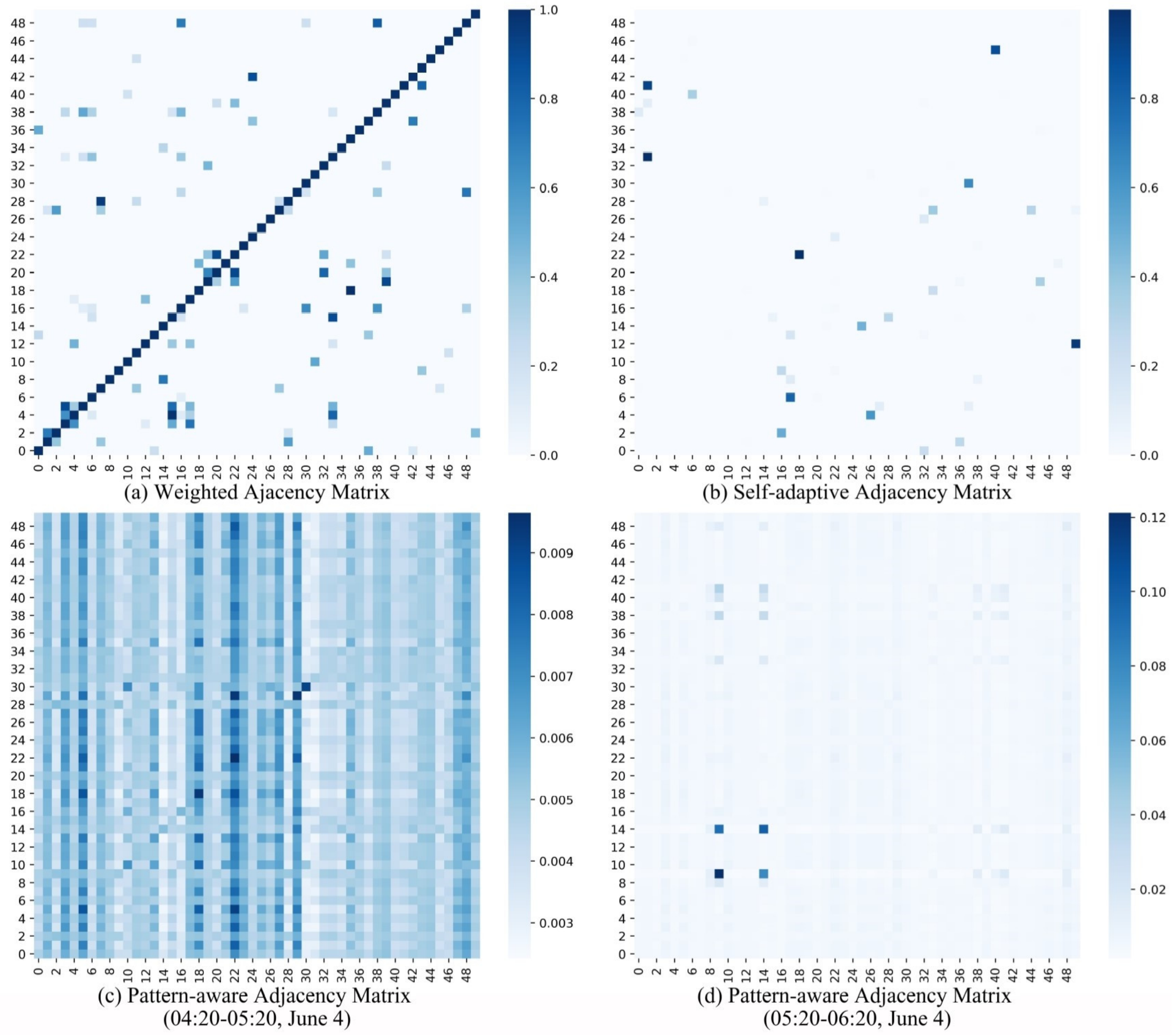

(a) Weighted Ajacency Matrix

(b) Self-adaptive Adjacency Matrix

(c) Pattern-aware Adjacency Matrix (04:20-05:20, June 4)

(d) Pattern-aware Adjacency Matrix (05:20-06:20, June 4)

**Fig. 11.** The heatmaps of different adjacency matrices for the first 50 sensors.

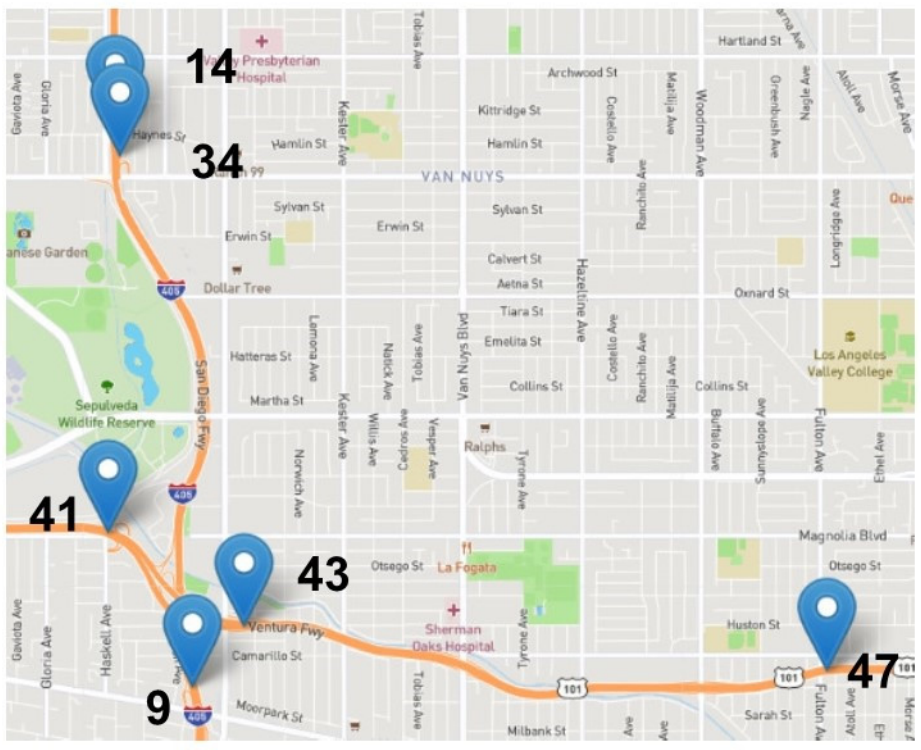

**Fig. 12.** Geographical locations of part of the sensors.

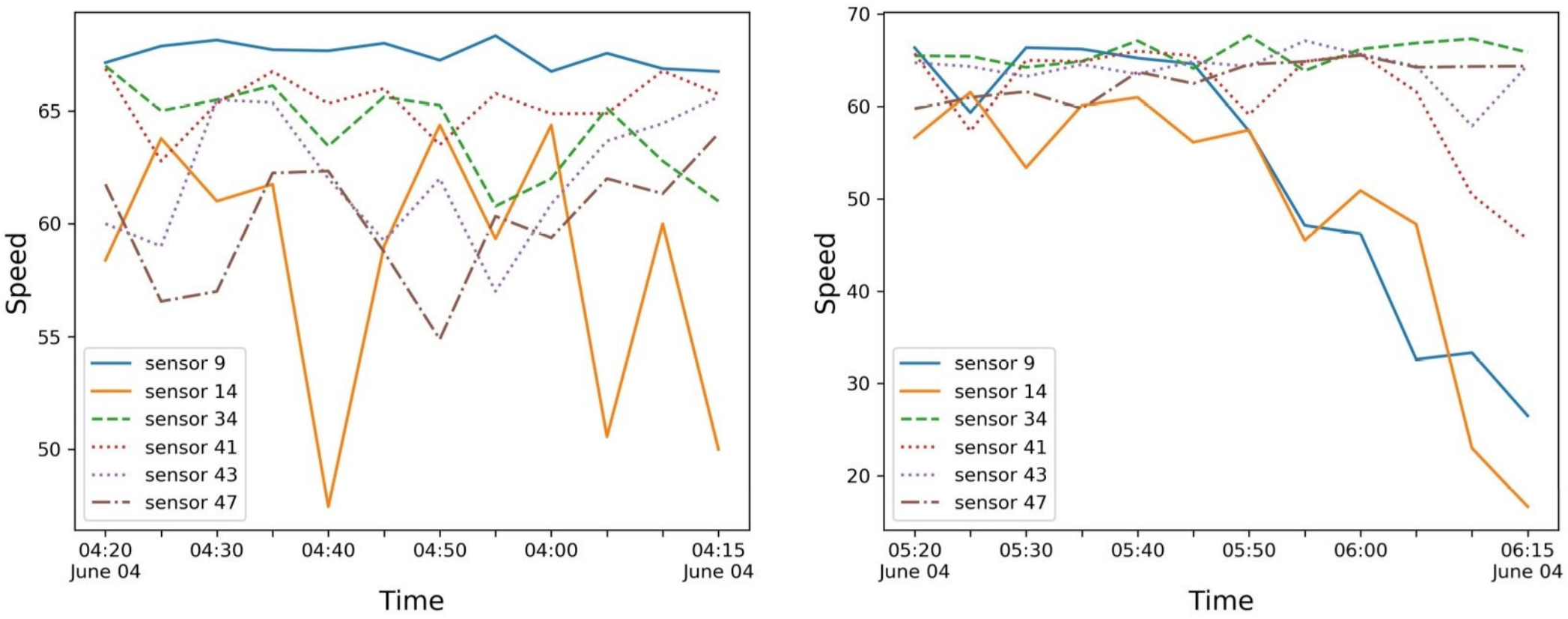

**Fig. 13.** The traffic time series of sensors during 04:20-05:20 (left), June 4 and 05:20-06:20, June 4 (right).

#### 4.3.4. Effect of temporal dependency modeling

To investigate the effect of temporal dependency modeling including the design of encoder-decoder architecture and the look-back attention mechanism, we further evaluate three variants of STSeq2Seq. *NoDec* removes the recurrent decoder, which means there is no decoding process and the output of projection layer is directly mapped to the prediction sequence. *RecEnc* replaces the encoder with DCGRU, which lead to the encoder being a recurrent architecture. *NoAttn* is similar to STSeq2Seq except that it has no look back attention mechanism used in the decoder. Fig. 14 shows the performance comparison of the above three variants and the complete STSeq2Seq model. We observe that STSeq2Seq consistently outperforms *RecEnc*, *NoDec*, and *NoAttn*, indicating the effectiveness of the proposed convolutional encoder, recurrent decoder, and look back attention mechanism when modeling complex temporal dependency. Moreover, STSeq2Seq's superiority in forecasting accuracy becomes more obvious as the predictions are made further into the future. We argue that this is because both global and local spatiotemporal features are learned during encoding phase and the dynamics of future time series are fully considered in the decoding phase. The look-back attention mechanism further enhances the model through explicit modeling of the dynamic temporal dependency between historical and future time steps.

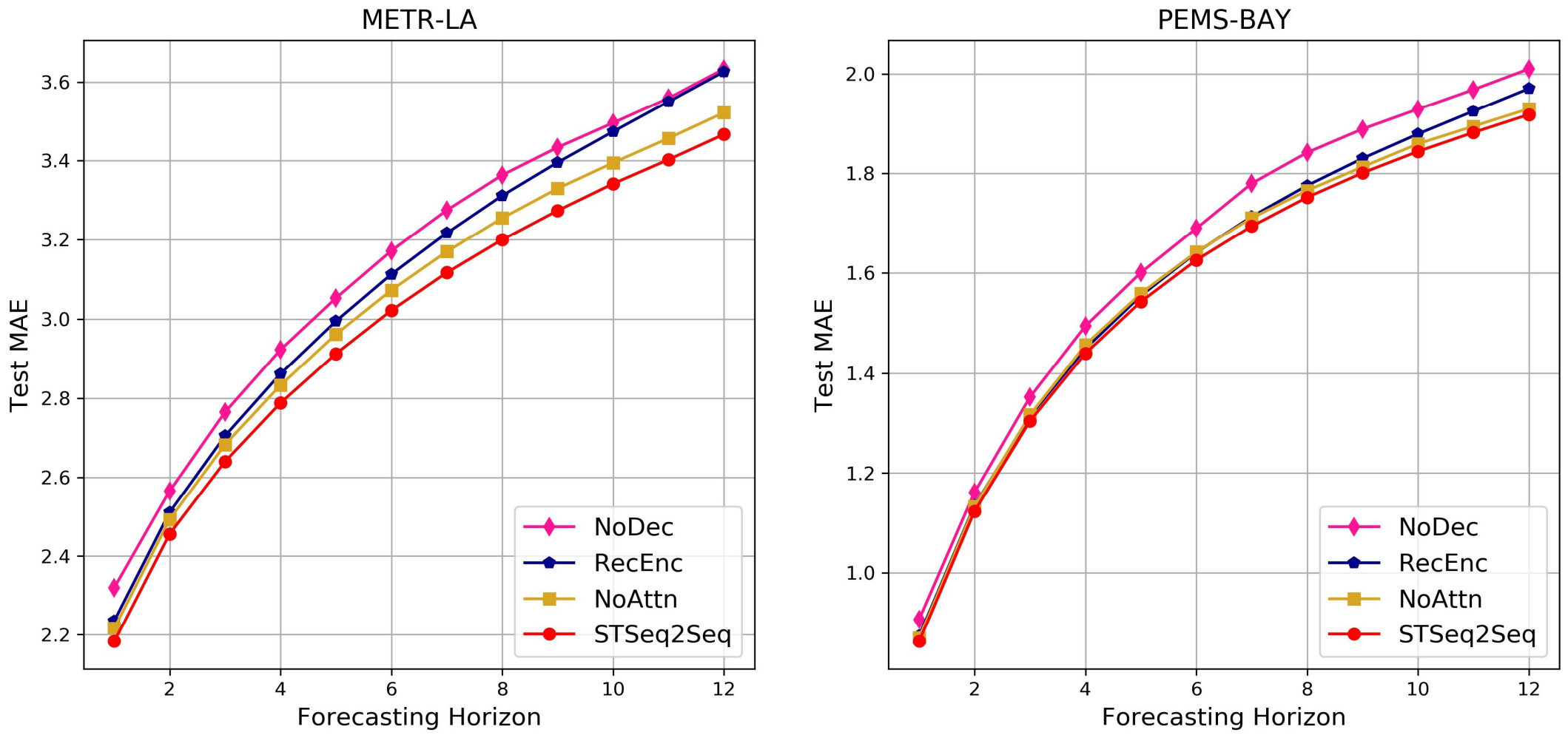

**Fig. 14.** Test MAE under different configurations of seq2seq architectures.

In Section 3.2.3, we formalized the look-back attention mechanism and suggested that the attention weight $\alpha_{ij}$ could enable the model to capture the dynamic temporal dependency by specifying how much the historical time steps $j$ should be attended to when making predictions at future time steps $i$. To get a deeper understanding of the proposed attention mechanism, a visualization of the heatmaps of attention coefficients is shown in Fig. 15. Specifically, the attention coefficients were calculated and averaged over one-hour interval when predicting on the test set of PEMS-BAY, and the results of four representative periods (i.e. morning non-rush hour 5am-6am, morning rush hour 8am-9am, afternoon non-rush hour 2pm-3pm, and evening rush hour 5pm-6pm) were selected for visualization. To reveal the traffic condition, the speed statistics of corresponding periods were presented in Fig. 16, where we drew frequency histograms of speed readings and fitted probability density functions to them using kernel density estimation (Parzen, 1962). We inspected the visualization results and made following observations:

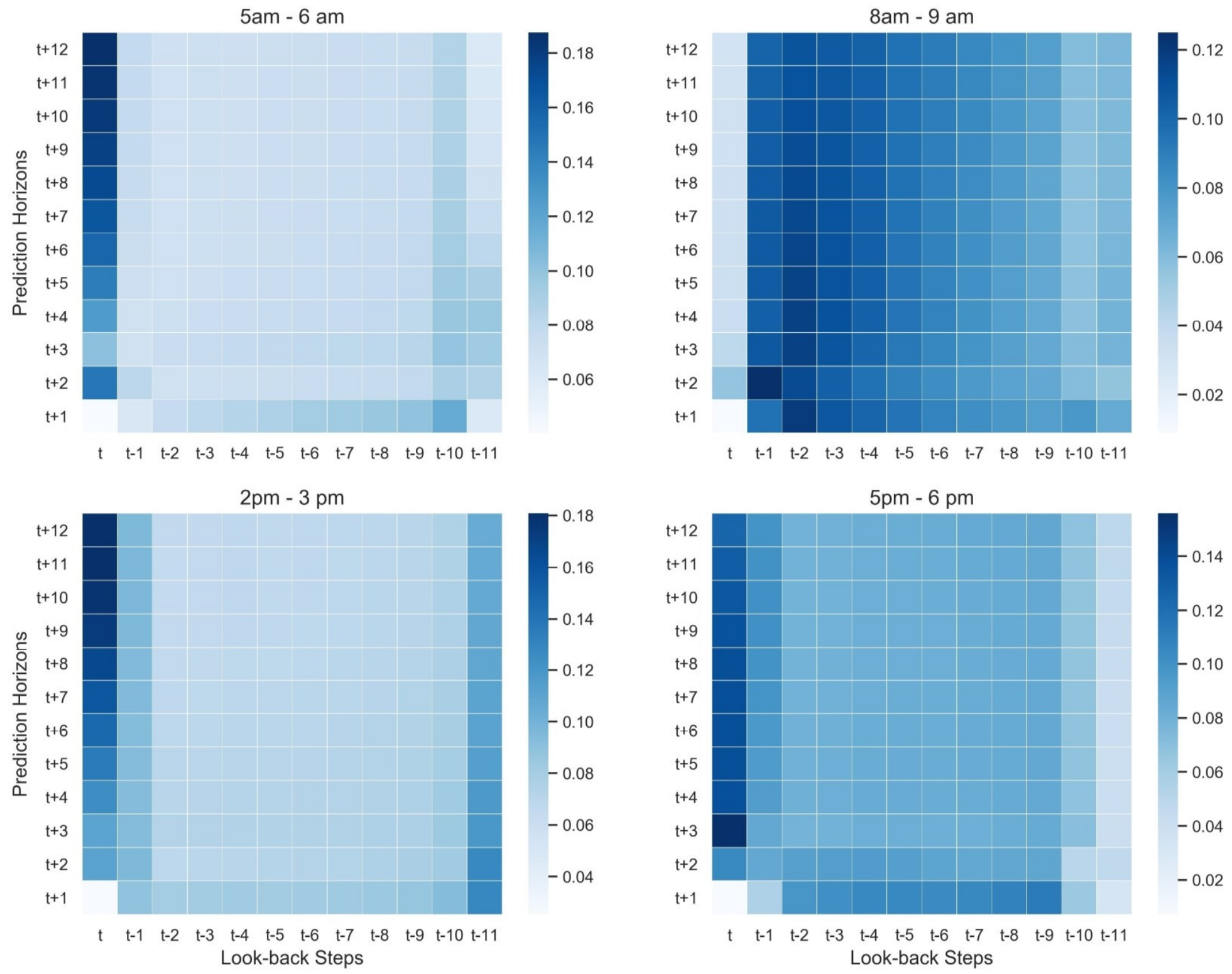

**Fig. 15.** The heatmaps of look-back attention weights (averaged over entire test set).

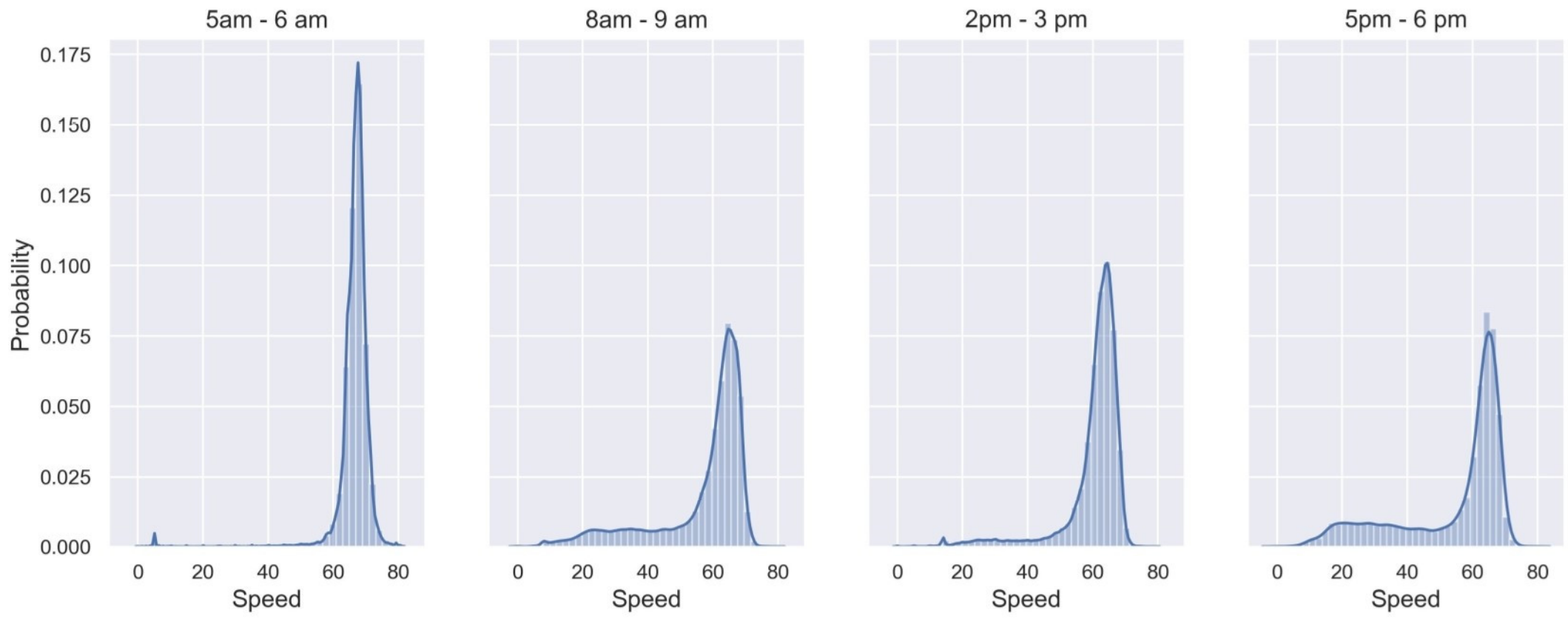

**Fig. 16.** Statistics of speed readings within four selected time intervals.

(1) The attention distribution shows similar patterns for periods 5am–6am and 2pm–3pm (non-rush hours), while it differs significantly for 8am-9am and 5pm–6pm (rush hours).

(2) The attention weights during rush hours (i.e. 8am-9am and 5pm–6pm) exhibit relatively uniform distributions over all look-back time steps compared to that of non-rush hours (i.e. 5am–6am and 2pm–3pm), in which the first look-back step takes up the most of the attention weights.

(3) The corresponding statistic information illustrated in Fig. 16 shows that the speed observations during 5am–6am and 2pm–3pm (non-rush hours) concentrate more on high values, indicating that the traffic is smooth and uncongested, whereas, the speed has larger variance during 8am–9am and 5pm–6pm (rush hours), which indicate relatively congested traffic conditions.

We concluded from above observations that for uncongested traffic, the attention scheme is comparatively simple and the model tends to pay attention to most recent information, while congested traffic causes complex attention distribution and the model consults wider range of look-back time steps accordingly. This concurs with the findings of (Ermagun and Levinson, 2019), in which the authors discovered that "the look-back time window for the uncongested traffic regime is shorter than the congested one".

#### 4.3.5. Model efficiency

Table 3 presents the computation time of DCRNN, Graph WaveNet and STSeq2Seq on the METR-LA dataset. We recorded the average time cost for each epoch during training and the total elapsed time for validation. It can be observed that during training, STSeq2Seq runs about two times faster than DCRNN due to the parallel computation of its convolutional encoder, but is about two times slower than Graph WaveNet because Graph WaveNet does not have a dedicated decoder. For inference, STSeq2Seq still achieves fairly efficient performance, since it can predict 6850*207 (the size of the test set) traffic speed values within only a few seconds. To summarize, STSeq2Seq reduces the computation overhead by a large margin comparing to traditional seq2seq architectures. But there is still room for further improvement.

**Table 3.** Comparison of computation time on the METR-LA dataset.

| Model | Computation Time | |
|---|---|---|
| | Training (s/epoch) | Inference (s/epoch) |
| DCRNN | 203.13 | 29.64 |
| Graph WaveNet | 46.35 | 2.07 |
| STSeq2Seq | 94.09 | 7.97 |

## 5. Conclusion

In this research, we focus on network-wide multiple steps ahead traffic forecasting. We proposed a novel deep learning framework named STSeq2Seq, in which a spatial convolution operation was defined by weighted adjacency matrix and pattern-aware adjacency matrix to explicitly model both local and dynamic non-local spatial correlation between traffic links, and a seq2seq architecture combining a convolutional encoder and a recurrent decoder was developed to account for temporal dependency modeling. Extensive experiments were conducted on two real-world publicly-available traffic datasets and the experiment results demonstrate state-of-the-art performance of the proposed model. Based on the empirical results, the noteworthy findings are stated as follows:

- The weighted adjacency matrix alone is very useful in most cases, especially under relatively stable traffic conditions.
- The proposed pattern-aware adjacency matrix is shown to be capable of learning temporal patterns of traffic series, and has the potential to capture dynamic non-local spatial correlations among traffic links.
- The proposed seq2seq architecture which facilitate hierarchical encoding and autoregressive decoding is effective in improving forecasting performance as well as reducing computation overhead.
- The proposed look-back attention mechanism for capturing non-stationary temporal dependency is effective. It improves forecasting accuracy and provides interpretation about the exploited temporal correlations.

Although we only validated the pattern-aware adjacency matrices for short-term traffic forecasting,

they could find potential applications in other network modeling problems such as spatial analysis in geographical information sciences and social network analysis in social sciences. The whole architecture of STSeq2Seq could also be applied to other spatiotemporal forecasting tasks such as whether, land use and air quality forecasting. The following suggestions are made for future research:

- Investigating the generalization ability of the proposed model for diverse traffic conditions (e.g. traffic during peak hours, on weekends and holidays, and during planned or accidental disruptions)
- To examine the effect of incorporating exogenous data (e.g. weather condition, speed limit on roads), should they be available.
- There are some recent studies that propose the complementary and competitive nature of traffic link (Ermagun et al., 2017) and expect a positive and a negative spatial dependency between complementary and competitive links, respectively (Ermagun and Levinson, 2018a). The present study does not explicitly address such an issue and the methodology we proposed for capturing spatial correlation is based on the similarity between traffic patterns. The idea of capturing the competitive links remains to be explored.
- Domain knowledge could be integrated into the proposed model. For example, utilize the propagation waves of traffic flow (Zhang et al., 2018) or apply queuing theory and simulate individual behaviors in traffic (Cascetta, 2013). Hopefully, these considerations could improve the forecasting ability as well as interpretability of the model.

## CRediT authorship contribution statement

**Xinglei Wang**: Conceptualization, Methodology, Software, Validation, Formal analysis, Investigation, Data curation, Writing - original draft, Writing – review & editing, Visualization. **Xuefeng Guan**: Conceptualization, Methodology, Validation, Resources, Writing - review & editing, Supervision, Project administration, Funding acquisition. **Jun Cao**: Investigation, Data curation, Writing - Review & Editing. **Na Zhang**: Investigation, Data curation, Writing – Review & Editing. **Huayi Wu**: Resources, Supervision, Project administration, Funding acquisition.

## Acknowledgements

This work is supported in part by the National Natural Science Foundation of China [Grant number 41971348] and the National Key Research Development Program of China [Grant number 2017YFB0503802].